\RequirePackage{fix-cm}
\documentclass{article}
\usepackage{arxiv}
\usepackage{graphicx}
\usepackage{balance} 
\usepackage{tikz}
\usepackage{pgf}
\usepackage{float}
\usepackage[utf8]{inputenc}
\usepackage{pgfplots}
\usepackage{pgfplots}
\usepgfplotslibrary{patchplots,colormaps}
\pgfplotsset{compat=1.9} 
\DeclareUnicodeCharacter{2212}{−}
\usepgfplotslibrary{groupplots,dateplot}
\usetikzlibrary{patterns,shapes.arrows}
\usepackage{graphicx}
\usepackage{caption}
\usepackage{subcaption}
\usepackage{array}
\usepackage{amsmath}
\usepackage{amssymb}
\usepackage{amsfonts}
\usepackage{bm}
\usepackage{xcolor}
\usepackage{multirow,array}
\usepackage{comment}
\usepackage{url} 
\usepackage{tikz}
\usepackage{pgfplots}
\usepackage[normalem]{ulem}
\usetikzlibrary{matrix,positioning}
\usepackage{makecell}
\usepackage[flushleft]{threeparttable}
\usepackage{multicol}
\usepackage{graphicx}
\usepackage[affil-it]{authblk}
\usetikzlibrary{automata,positioning,arrows.meta,math,external}
\usetikzlibrary{decorations.pathreplacing}
\usetikzlibrary{shapes,shapes.geometric, snakes}
\usetikzlibrary{arrows, chains, fit, quotes}
\usepackage[eulergreek]{sansmath}
\usepackage[hidelinks]{hyperref}
\usepackage{arxiv} 

\pgfmathdeclarefunction{norm}{3}{%
  \pgfmathparse{sqrt(0.5*#3/pi)*exp(-0.5*#3*(#1-#2)^2)}%
}

\usepackage{pgfplots}
\usetikzlibrary{arrows.meta}

\pgfplotsset{compat=newest,
    width=8.5cm,
    height=7cm,
    scale only axis=true,
    max space between ticks=25pt,
    try min ticks=5
}
\tikzset{
    semithick/.style={line width=0.8pt},
}
\usepgfplotslibrary{groupplots}
\usepgfplotslibrary{dateplot}

\usepackage[ruled,linesnumbered,lined,commentsnumbered]{algorithm2e}
\usepackage{cancel}
\usepackage{eucal}
\usepackage{amssymb}
\usepackage{marvosym}
\begin{document}


\title{Monte Carlo Tree Search Algorithms for Risk-Aware and Multi-Objective Reinforcement Learning \thanks{An earlier version of this work was presented as an extended abstract at the \emph{International Conference of Autonomous Agents and Multi-Agent Systems} 2021 \cite{hayes2021dmcts} and at the \emph{Adaptive and Learning Agents Workshop} 2021 \cite{hayes2021dmcts_long}. This article extends our previous work with additional theoretical analysis and new empirical results.}}


\author[1, \Letter]{\textbf{Conor F. Hayes}}
\author[2]{\textbf{Mathieu Reymond}}
\author[2, 3]{\textbf{Diederik M. Roijers}}
\author[1]{\textbf{Enda Howley}}
\author[1]{\textbf{Patrick Mannion}}
\affil[1]{University of Galway, Ireland}
\affil[2]{Vrije Universiteit Brussel, Belgium}
\affil[3]{City of Amsterdam, The Netherlands}
\affil[\Letter]{c.hayes13@nuigalway.ie}



\maketitle

\begin{abstract}
In many risk-aware and multi-objective reinforcement learning settings, the utility of the user is derived from a single execution of a policy. In these settings, making decisions based on the average future returns is not suitable. For example, in a medical setting a patient may only have one opportunity to treat their illness. Making decisions using just the expected future returns -- known in reinforcement learning as the value -- cannot account for the potential range of adverse or positive outcomes a decision may have. Therefore, we should use the distribution over expected future returns differently to represent the critical information that the agent requires at decision time by taking both the future and accrued returns into consideration. 
In this paper, we propose two novel Monte Carlo tree search algorithms. Firstly, we present a Monte Carlo tree search algorithm that can compute policies for nonlinear utility functions (NLU-MCTS) by optimising the utility of the different possible returns attainable from individual policy executions, resulting in good policies for both risk-aware and multi-objective settings. Secondly, we propose a distributional Monte Carlo tree search algorithm (DMCTS) which extends NLU-MCTS. DMCTS computes an approximate posterior distribution over the utility of the returns, and utilises Thompson sampling during planning to compute policies in risk-aware and multi-objective settings. Both algorithms outperform the state-of-the-art in multi-objective reinforcement learning for the expected utility of the returns.  \keywords{Multi-objective; risk-aware; decision making; distributional; reinforcement learning; Monte Carlo tree search}

\end{abstract}

\section{Introduction}
In real-world decision making, a policy is often only executed once. For example, consider a government planning to build an off-shore wind farm to generate electricity. To ensure electricity generation is maximised, the wind farm must be located in an area with sufficient wind while also not interfering with any fishing routes or protected marine life. Given the off-shore wind farm will only be constructed once, the government must consider each potential outcome and likelihood of each outcome to ensure an optimal decision can be made. 


In reinforcement learning (RL), the expected return is used to make decisions \cite{sutton1998introduction}. However, in many scenarios the utility of a user is derived from a single execution of a policy, and, therefore, the utility of the returns must be optimised \cite{roijers2018multi}. For example, in a medical setting a patient may only have one opportunity to select a treatment. In this example, a patient will aim to cure their illness based on a single course of a treatment. As such, the user's utility is derived from the single execution of a policy. Moreover, computing a policy based on applying a utility function to expected return is incompatible with how the user's utility is derived because the expected return considers the average outcome over multiple policy executions. Therefore, the utility of the expectation is computed. In contrast, a policy that maximises utility of the return considers the utility obtained from each individual outcome, which is compatible with how the user's utility is derived. Therefore, the expected utility must be maximised (see Section \ref{sec:ser_vs_esr}). 

When optimising for expected utility the underlying distribution of the returns must be used differently. Therefore, decisions must be made using the utility of the returns of a full policy. Under these conditions, an agent must be able to sample from the underlying return distribution to calculate the future returns. The agent must also be able to calculate the returns accrued at each timestep. Therefore, we theorise that for an agent to have sufficient critical information at decision time the agent must apply the utility function to the cumulative returns, which is the sum of the accrued and future returns \cite{roijers2018multi}.

To calculate the utility, we apply the utility function to the returns where a user's utility function is known a priori. In other words, in the taxonomy of multi-objective sequential decision making \cite{roijers2013survey}, we are in the known utility function scenario. When optimising under the expected utility, it is critical to only apply the utility function to the returns of a full execution of a policy \cite{roijers2018multi} because nonlinear utility functions do not distribute across the sum of immediate and future returns \cite{roijers2018multi,hayes2021practical}. In this case, the agent must know the returns it has already accrued and the future returns before applying the utility function. For example, before the $2008$ financial crash, many investment bankers were guaranteed their base salaries regardless of their losses, but their bonuses were dependent on their returns from investments. In the case of an investor incurring a loss, the only policy that would result in a bonus would be one that executes an increasingly risky strategy to win back the losses and receive some bonus.

Learning the utility of the returns is thus naturally risk-aware. Optimising the utility of the sum of the accrued and future returns to make decisions enables an agent to avoid certain undesirable outcomes. Without knowing the accrued returns, an agent cannot understand how future actions could affect the cumulative return. To make decisions that maximise the user's utility, the agent must have information about both the accrued and the future returns.
 
A further complicating factor is that, in the real world, decision making often involves trade-offs based on multiple conflicting objectives \cite{dulac-arnold2021,vamplew2021scalar,reymond2022exploring}. For example, we may want to maximise the power output of coal-burning electrical generators while minimising $CO_{2}$ emissions. Many approaches to multi-objective decision making only consider linear utility functions; this limitation severely restricts the real-world applicability of these methods \cite{vamplew2008limitations}, given that utility in many real-world problems is derived in a nonlinear manner.
 
In the multi-objective case, optimising under the expected utility is known as optimising the expected scalarised returns (ESR) criterion. For multi-objective reinforcement learning (MORL), the utility function expresses the user's preferences over objectives. If the utility function is linear and is known a priori, it is possible to translate a multi-objective decision problem to its single-objective equivalent. Once translated, we can then apply single objective methods to solve the decision problem. However, if the utility function is nonlinear, as human preferences often are, explicitly multi-objective methods are required to find optimal solutions \cite{roijers2013survey}. The majority of MORL algorithms focus on the scalarised expected returns (SER) criterion. It has been shown that for nonlinear utility functions the policies learned under the ESR criterion and the SER criterion can be different \cite{Radulescu2019Equilibria}. nonlinear utility functions invalidate the Bellman equation, given nonlinear utility functions do not distribute across the sum of the immediate and future returns, which restricts the number of usable algorithms in this setting \cite{hayes2021practical}. Therefore, to increase MORL's usability in the real world, dedicated multi-objective algorithms for the ESR criterion and the SER criterion that can learn policies for nonlinear utility functions must be formulated. We note that the MORL literature focuses almost exclusively on the SER criterion, leaving the ESR criterion largely understudied with a few exceptions \cite{roijers2018multi,hayes2021esr_set,hayes2021expected,malerba2021esr,vamplew2021stocashtic,hayes2022decision,hayes2022modvi_esr}\footnote{We will make the distinction between the SER and ESR criterion clear in a later section}. 

We propose a novel algorithm that can optimise for nonlinear utility functions for expected utility by taking both the accrued and future returns into consideration. To do so, we define a nonlinear utility function Monte Carlo tree search (NLU-MCTS) algorithm that performs Monte Carlo roll-outs to calculate the future returns while also calculating the accrued returns. Therefore, NLU-MCTS can make decisions using the expected utility of the returns over multiple policies. NLU-MCTS builds upon Monte Carlo tree search and uses UCB to explore during planning.

In sequential decision making settings, one of the fundamental challenges is the exploration versus exploration dilemma \cite{sutton1998introduction}. Thompson sampling is an algorithm that has been shown to address this dilemma in bandit settings \cite{chapelle2011thompson}. Thompson sampling selects actions based on the probability matching principle, where actions are selected stochastically based on the probability of the action being optimal. Thompson sampling has been shown to empirically outperform UCB in bandit settings, and under a wide range of problems shows a more robust convergence compared to UCB \cite{bai2014thompson}. Therefore, to exploit the potential performance gains of Thompson sampling we propose a new algorithm known as distributional Monte Carlo tree search (DMCTS) which computes an approximate posterior distribution over the expected utility over the returns. Using the computed approximate posterior distribution it is possible to use Thompson sampling methods to explore during planning.

Both NLU-MCTS and DMCTS overcome the issues present when making decisions solely with the expected return \cite{browne2012mcts,silver2016,Sebag2012,Kocsis2006,Painter2020}. As we will show, computing the utility of the returns of a policy is useful when optimising for risk-aware RL and under the MORL ESR criterion, given utility of the returns of a policy contains more information about the range of potential negative and positive outcomes during planning and at decision time. NLU-MCTS achieves good performance in both risk-aware and multi-objective settings, while DMCTS achieves good performance in risk-aware settings and state-of-the-art performance under multi-objective ESR settings.


\section{Background}
\label{sec:background}
In this section we introduce necessary background material, including multi-objective Markov decision processes, the known utility function scenario, commonly used optimality criteria in multi-objective decision making, risk-aware utility functions, Bootstrap Thompson Sampling, Expected Utility Policy Gradient and Monte Carlo tree search.
\subsection{Multi-Objective Reinforcement Learning}
In multi-objective reinforcement learning (MORL), we deal with decision problems with multiple objectives \cite{roijers2013survey,mannion2021multi}, often modelled as a multi-objective Markov decision process (MOMDP). A MOMDP is a tuple, $\mathcal{M} = (\mathcal{S}, \mathcal{A}, \mathcal{T}, \gamma, \mathcal{R})$, where $\mathcal{S}$ and $\mathcal{A}$ are the state and action spaces, $\mathcal{T} \colon \mathcal{S} \times \mathcal{A} \times \mathcal{S} \to \left[ 0, 1 \right]$ is a probabilistic transition function, $\gamma$ is a discount factor determining the relative importance of future rewards and $\mathcal{R} \colon \mathcal{S} \times \mathcal{A} \times \mathcal{S} \to \mathbb{R}^n$ is an $n$-dimensional vector-valued immediate reward function. In MORL, $n>1$.

\subsection{The Known Utility Function Scenario}
\begin{figure}
\centering
\begin{tikzpicture}[>={Stealth[width=6pt,length=9pt]}, skip/.style={draw=none}, shorten >=1pt, accepting/.style={inner sep=1pt}, auto]
\draw (-20.0pt,20.0pt) node[below](0) {\begin{tabular}{c} MOMDP \\ + \\ utility function \end{tabular}} ;
\draw (75.0pt, 0.0pt) node[rounded rectangle, thick, fill=gray!20, minimum height=1cm,minimum width=2cm, draw, label=below:$\text{\emph{planning or learning phase}}$](1){$\text{algorithm}$};
\draw (240.0pt,14.5pt) node[below, minimum height=1cm,minimum width=2cm, label=below:$\text{\emph{execution phase}}$](2){$\text{single solution}$} ;
\draw [dashed] (130.0pt,20pt) -- (130.0pt,-20.0pt);
\path[->] (0) edge node{} (1);
\path[->] (1) edge node{} (2);
\end{tikzpicture}
\caption{The known utility function scenario \cite{hayes2021practical}.}
\label{fig:known_utility_function_scenario}
\end{figure}
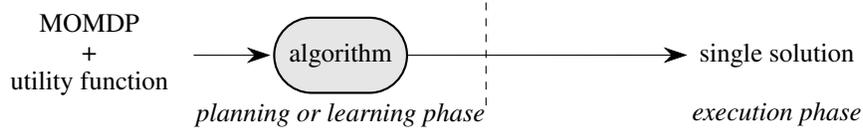

In MORL, an agent seeks to maximise a user's utility function, where a user's utility function describes their preferences over objectives. In certain scenarios the utility function of a user can be known at the time of learning or planning. In the taxonomy of MORL we are deemed to be in the known utility function scenario \cite{hayes2021practical,roijers2013survey}. When the utility function of a user is known, a single optimal policy can be computed. Figure \ref{fig:known_utility_function_scenario} describes the phases of the known utility function scenario. There are two phases in the known utility function scenario: the planning or learning phase and the execution phase. During the planning or learning phase a multi-objective reinforcement learning or planning algorithm is deployed in the MOMDP to compute a single optimal policy for the known utility function. A single optimal policy is computed once the algorithm has completed planning or learning. The computed policy is then executed during the execution phase.

\subsection{Risk-Aware Utility Functions}
\label{sec:risk_aware_utility_functions}
In single-objective decision making under uncertainty, utility functions are often utilised \cite{von1947theory}. In scenarios where risk is considered, utility functions are often used to represent a user's preference for risk. In risk-aware settings the utility function is applied to the returns and the expected utility is maximised. When making decisions in scenarios with risk, a user can be described as risk-averse, risk-seeking or risk-neutral. 

A user's preference for risk can be described by the shape of their utility function \cite{friedman1948,arrow1965}. The shape of a risk-seeking utility function is convex. For example the nonlinear utility function $u(x) = x^{2}$ is a risk-seeking utility function given its shape is convex \cite{machina1987utility,friedman1948}. For a user that is risk-averse their utility function is concave. For example the nonlinear utility function $u(x) = x^{0.5}$ is risk-averse given the utility function has a concave shape from below \cite{karni1991utility,friedman1948}. In contrast to risk-averse and risk-seeking utility functions, risk-neutral utility functions are linear \cite{karni1991utility}. A user who has a risk-neutral utility function has no preferences for risk and therefore the utility is a linear function of the returns. For example, $u(x) = x$ is a risk-neutral and linear utility function.  In this paper we only consider nonlinear utility functions, therefore we focus on risk-seeking and risk-averse utility functions.



\subsection{Scalarised Expected Returns vs Expected Scalarised Returns}
\label{sec:ser_vs_esr}
In MORL, the user's utility derives from the vector-valued outcomes (returns). This is typically modelled as a utility function that needs to be applied to these outcomes in one way or another. For this, we consider two choices \cite{roijers2013survey,hayes2021practical}. Calculating the expected value of the return of a policy before applying the utility function leads to the scalarised expected returns (SER) optimisation criterion: 
\begin{equation}
    V_{u}^{\pi} = u\left(\mathbb{E} \left[ \sum\limits^\infty_{t=0} \gamma^t {\bf r}_t \:|\: \pi, \mu_0 \right]\right).
    \label{eqn:ser}
\end{equation}
SER is the most commonly used criterion in the multi-objective (single agent) planning and reinforcement learning literature \cite{roijers2013survey}. For SER, a coverage set is defined as a set of optimal policies for all possible utility functions. 

In contrast to the SER criterion, if the utility function is applied before computing the expectation, then the expected scalarised returns (ESR) criterion is being optimised \cite{roijers2018multi}: 
\begin{equation}
    V_{u}^{\pi} = \mathbb{E} \left[ u\left( \sum\limits^\infty_{t=0} \gamma^t {\bf r}_t \right) \:|\: \pi, \mu_0 \right].
    \label{eqn:esr}
\end{equation}
Similar to risk-aware settings for single objectives (see Section \ref{sec:risk_aware_utility_functions}), the ESR criterion maximises the expected utility. Therefore, the ESR criterion is naturally risk-aware while considering multiple objectives. ESR is the most commonly used criterion in the game theory literature on multi-objective games~\cite{radulescu2020survey}, with some exceptions (e.g. \cite{Radulescu2019Equilibria}).

\subsection{Monte Carlo Tree Search}
One way of approaching a decision problem is to use tree search. Perhaps the most popular of such methods is
Monte Carlo tree search (MCTS) \cite{Coulom2006}, which employs heuristic exploration to construct its search tree. MCTS builds a search tree of nodes, where each node has a number of children. Each child node corresponds to an action available to the agent. MCTS has two phases: the planning phase and the execution phase.

In the planning phase the agent implements the following four steps \cite{browne2012mcts}: selection, expansion, simulation and backpropagation. \textbf{Selection:} the agent traverses the search tree until it reaches a node for which not all of its possible child nodes have been explored. \textbf{Expansion:} at a node whose children have not all been expanded, the node must be expanded. The agent creates a random child node and then must simulate the environment for the newly created child node. \textbf{Simulation:} the agent executes a random policy through Monte Carlo simulations until a terminal state of the environment is reached. The agent then receives the returns. \textbf{Backpropagation:} the agent must backpropagate the returns received at a terminal state to each node visited during selection where a predefined algorithm statistic e.g. UCB \cite{Coulom2006,Kocsis2006} is updated. Each step is repeated a specified number of times, which incrementally builds the search tree. Or, as we will discuss in the next subsection, a posterior belief on the returns, from which we can draw actions using Thompson sampling \cite{Bai2014}.

During the execution phase the agent must select a child node, corresponding to an action and associated state transition, to traverse to next. The agent evaluates the statistic at each node that is reachable from the root node and moves to the node which returns the maximum value. Once the execution phase has completed, the agent repeats the planning phase.

As already highlighted MCTS makes decisions and explores based on a predefined algorithm statistic. One such version of MCTS is UCT \cite{Kocsis2006} which uses the following formula to derive the optimal action at decision time while also incorporating exploration during learning:
\begin{equation}
    v_{i} + C \times \sqrt{\frac{ln(N)}{n_{i}}},
    \label{eqn:UCT}
\end{equation}
where $v_{i}$ is the approximated value of the node $i$, $n_{i}$ is the number of the times the node $i$ has been visited and $N$ is the total number of times that the parent of node $i$ has been visited. $C$ is a hyperparameter that can be tuned for exploration, however $C$ is often set to $\sqrt{2}$.

\label{sec:MCTS}

\subsection{(Bootstrap) Thompson Sampling}
\label{sec:BTS}
As previously mentioned, during the planning phase of MCTS, we can use Thompson sampling to take exploring actions \cite{Bai2014}. However, it is not always possible to get an exact posterior. 
In this case a bootstrap distribution over means can be used to approximate a posterior distribution \cite{efron2012,newton1995}. Eckles et al.\ \cite{EcklesKaptein2014,EckelsKaptein2019} use a bootstrap distribution to replace the posterior distribution used in Thompson Sampling. This method is known as Bootstrap Thompson Sampling (BTS) \cite{EcklesKaptein2014} and was proposed in the multi-arm bandit setting. The bootstrap distribution contains a number of bootstrap replicates, $j \in  \{1, ..., J\}$, where $J$ is a hyper-parameter that can be tuned for exploration. For a small $J$, BTS can become greedy. A larger $J$ value increases exploration,  but at a computational cost \cite{EcklesKaptein2014}. 

Each bootstrap replicate, $j$, in the bootstrap distribution contains two parameters, $\alpha_{j}$ and $\beta_{j}$, where $\frac{\alpha_{j}}{\beta_{j}}$ is an is an estimate of replicate $j$'s expected utility. At decision time, to determine the optimal action the bootstrap distribution for each arm, $i$, is sampled. The observation for the corresponding bootstrap replicate, $j$, is retrieved and the arm with the maximum expected utility is pulled \cite{EcklesKaptein2014}.  

The distribution which corresponds to the maximum arm is randomly re-weighted by simulating a coin-flip (commonly known as sampling from a Bernoulli bandit) for each bootstrap replicate, $j$, in the bootstrap distribution (see Algorithm \ref{alg:background_bts_update}). If the coin-flip is heads, the $\alpha$ and $\beta$ parameters for $j$ is re-weighted\footnote{Updating the distribution in this way is known as "double-or-nothing" or online half sampling \cite{EcklesKaptein2014}. It is important to note that the absolute scale of the weights does not matter for most estimators \cite{EcklesKaptein2014}. In the literature various other weight distributions have been used. For example, Rubin \cite{rubin1981} uses a Bayesian bootstrap which uses exponential weights. While this overcomes some numerical problems, it requires updating all replicates and therefore can be more computationally expensive. For an extensive study on weight distributions for bootstrapping the sample mean see Owen and Eckles \cite{owen2012bootstrapping}.}. To do so, the return is added to the $\alpha_{j}$ value and $1$ is added to $\beta_{j}$ \cite{EcklesKaptein2014}. 

\begin{algorithm}
\For{$j \in J$}
{\text{sample } $d_{j}$ \text{from Bernoulli(1/2)} \\
\If{$d_{j} = 1$}{
$\alpha_{j} \leftarrow \alpha_{j} + u(\textbf{r})$\\
$\beta_{j} \leftarrow \beta_{j} + 1$\\}
}
\caption{Bootstrap Thompson Sampling Update}
\label{alg:background_bts_update}
\end{algorithm}

Bootstrap methods with random re-weighting \cite{rubin1981} are more computationally appealing as they can be conducted online rather than re-sampling data \cite{Oza2005}. BTS addresses problems of scalability and robustness when compared to Thompson Sampling \cite{EcklesKaptein2014}. Furthermore, bootstrap distributions can approximate posteriors that are difficult to represent exactly.

\subsection{Expected Utility Policy Gradient}
We now introduce Expected Utility Policy Gradient (EUPG) \cite{roijers2018multi}, a state-of-the-art MORL algorithm for ESR that we will use as a benchmark algorithm in our experiments. EUPG is an extension of Policy Gradient \cite{Sutton1999,williams1992}, where Monte Carlo simulations are used to compute the returns and optimise the policy. EUPG calculates the accrued returns, $\textbf{R}^-_{t}$, which is the sum of the immediate returns received as far as the current timestep, $t$. EUPG also calculates the future returns, $\textbf{R}^+_{t}$, which is the sum of the immediate returns from the current timestep, $t$, to the terminal state. Using both the accrued and future returns enables EUPG to optimise over the utility of the full returns of an episode, where the utility function is applied to the sum of $\textbf{R}^-_{t}$ and $\textbf{R}^+_{t}$. 

In policy gradient the policies are adapted towards the attained utility by gradient descent. For EUPG the utility of the sum of the accrued and future returns is calculated inside the loss function, which results in the following:
\begin{equation}
\label{eq:nll-utility}
    \mathcal{L}(\pi) = -\sum_{t=0}^{T} u(\textbf{R}^-_{t} + \textbf{R}^+_{t}) \log(\pi_\theta(a|s, \textbf{R}^-_{t}, t)).
\end{equation}
Roijers et al.\ \cite{roijers2018multi} demonstrated for the ESR criterion the accrued and future returns must be considered when learning in order to learn a good policy. Applying this consideration to EUPG, the algorithm achieves the state-of-the-art performance under the ESR criterion. In this paper, we use the same method of adding past and future returns together before applying the utility inside of the search scheme of our novel DMCTS algorithm. 

\section{Expected Scalarised Returns}
In the known utility functions scenario, there are two phases: the planning phase and the execution phase. During the learning phase a policy is computed and returned to the user. After planning has completed the user executes the computed policy during the execution phase. In scenarios where the utility of a user is derived from the single execution of a policy, the expected scalarised returns (ESR) criterion must be optimised. Under the ESR criterion, the user will only execute the computed policy once in the execution phase.\footnote{It is important to note that in order to compute policies during the planning phase multiple policy executions must take place. However, it is important to remember that the computed policy may then be executed once (ESR) or multiple times (SER) during the execution phase} The majority of RL research focuses on the SER criterion \cite{perny2010finding,pan2020additional}, while the ESR criterion has been largely overlooked with some exceptions \cite{roijers2018multi,hayes2021esr_set,malerba2021esr,vamplew2021stocashtic,hayes2022multi}. Additionally, the majority of RL research only considers linear utility functions. However, in the real world, utility functions can be nonlinear. A potential reason why the RL community has focused on linear utility function is that nonlinear utility functions invalidate the Bellman equation, given nonlinear utility functions do not distribute across the sum of the immediate and future returns \cite{roijers2018multi},
\begin{equation}
\label{eqn:nonlinear_distribute_sum}
\begin{split}
\max_\pi\ &\mathbb{E} \left[u\left({\bf R}_t^- + \sum_{i=t}^{\infty} \gamma^i {\bf r}_i\right)\ \middle|\ \pi, s_t \right] \not= \\
&u({\bf R}_t^-) + \max_\pi \mathbb{E}\left[  u\left( \sum_{i=t}^{\infty} \gamma^i {\bf r}_i\right)\ \middle|\ \pi, s_t \right],
\end{split}
\end{equation}
where $u$ is a nonlinear utility function\footnote{For nonlinear utility functions the accrued returns must be included in the state, $s_{t}$ \cite{roijers2018multi}.} and $\bf{R^{-}_{t}}$ $=$ $\sum_{i=0}^{t - 1} \gamma^i {\bf r}_i$. It has also been shown that for nonlinear utility functions the policies computed under the ESR criterion and the SER criterion can be different \cite{radulescu2020survey}. Therefore, to enhance RL's usability in real-world problem domains, new methods must be formulated that can compute policies for the ESR criterion and the SER criterion for nonlinear utility functions. 

When making decisions under the SER criterion, the expected returns is computed before the utility function is applied. A decision is then selected based on the scalar utility of the expectation \cite{vanmoffaert2013scalarised}. SER methods learn policies that optimise a user's utility function over multiple policy executions. Therefore, a user will execute a policy computed under the SER criterion multiple times during the execution phase. Under the SER criterion, making decisions on an expected value vector is optimal \cite{roijers2013survey}. However, a user optimising for the ESR criterion may only have one opportunity to execute a policy.  Therefore, making decisions based on a single expected value vector is not sufficient because the user must have sufficient critical information available about each potential return vector and the associated likelihood \cite{hayes2021esr_set,hayes2021expected}. Therefore, applying the utility function to each return vector before computing the expectation ensures  the user has taken into consideration each potential outcome a policy may have and the associated utility. We outline two new algorithms that compute policies under the ESR criterion in Section \ref{sec:nonlinear_MCTS} and Section \ref{sec:DMCTS}.


\section{Monte Carlo Tree Search for Nonlinear Utility Functions}
\label{sec:nonlinear_MCTS}
To compute policies for the ESR criterion when the utility function is nonlinear and known a priori \cite{hayes2021practical}, we present a novel Monte Carlo tree search algorithm, known as NLU-MCTS. As shown by Roijers et al. \cite{roijers2018multi}, in order to compute optimal policies under the ESR criterion, both the accrued and future returns must be taken into consideration before applying the utility function. Therefore, an algorithm must either maintain a distribution over the returns or have some method which allows the agent to sample from the underlying return distribution of the environment. NLU-MCTS utilises the latter, by performing Monte Carlo simulations to compute the future returns. Usually in single objective MCTS an expectation of the returns is maintained at each chance node and the agent seeks to maximise the expectation. When the utility function is nonlinear, making decisions based on the expected returns does not account for the potential undesired outcomes a decision may have. For risk-aware RL and MORL under the ESR criterion, we need to be able to make decisions with sufficient information to avoid undesirable outcomes and exploit positive outcomes. Our key insight is that computing the utility of the cumulative returns, the returns received from executing a policy, can be used to replace the expected future returns (of vanilla MCTS) at each node. We outline our algorithm for single-objective risk-aware RL and MORL that can compute policies for the ESR criterion.

\begin{figure}
\centering
\begin{tikzpicture}[>={Stealth[width=6pt,length=9pt]}, skip/.style={draw=none}, shorten >=1pt, accepting/.style={inner sep=1pt}, auto]
  \draw (0.0pt, 0.0pt)node[circle, fill=blue!20, thick, minimum height=0.6cm,minimum width=0.6cm, draw](0){};
   \draw (-40.0pt, -35.0pt)node[regular polygon,regular polygon sides=8, thick, fill=blue!20, minimum height=0.6cm,minimum width=0.6cm, draw](1){};
   \draw (40.0pt, -35.0pt)node[regular polygon,regular polygon sides=8, thick, fill=blue!20, minimum height=0.6cm,minimum width=0.6cm, draw](2){};
   \draw (-15.0pt, -70.0pt)node[circle, fill=blue!20, thick, minimum height=0.6cm,minimum width=0.6cm, draw](3){};
   \draw (-65.0pt, -70.0pt)node[circle, fill=blue!20, thick, minimum height=0.6cm,minimum width=0.6cm, draw](4){};
    \draw (15.0pt, -70.0pt)node[circle, fill=blue!20, thick, minimum height=0.6cm,minimum width=0.6cm, draw](5){};
   \draw (65.0pt, -70.0pt)node[circle, fill=blue!20, thick, minimum height=0.6cm,minimum width=0.6cm, draw](6){};

  \path[->] (0) edge node{} (1);
  \path[->] (1) edge node{} (3);
  \path[->] (1) edge node{} (4);
  
  \path[->] (0) edge node{} (2);
  \path[->] (2) edge node{} (5);
  \path[->] (2) edge node{} (6);


\end{tikzpicture}
\caption{A representation of a search tree constructed using NLU-MCTS for a problem with stochastic rewards and two actions. The search contains both decision nodes, represented by circular nodes, and chance nodes, represented by octagons.}
\label{fig:expectimax_dmcts}
\end{figure}
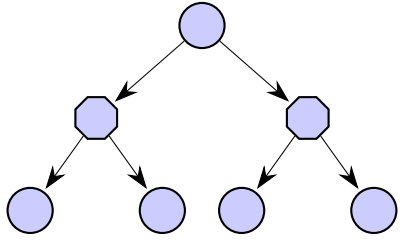

Before we outline how the accrued and future returns are computed, we must describe the structure of the search tree constructed by NLU-MCTS. Under the ESR criterion, the environment must be stochastic, where the state transitions or reward function are stochastic. To handle this uncertainty, NLU-MCTS builds an expectimax search tree using the same planning phase as MCTS (see Section \ref{sec:MCTS}). A search tree is a representation of the state-action space that is incrementally built via the steps of the underlying MCTS algorithm. An expectimax search tree \cite{Veness2011} uses both decision and chance nodes. Figure \ref{fig:expectimax_dmcts} describes a search tree constructed by NLU-MCTS which contains both decision and chance nodes. Each decision node represents a state, action and reward of a MOMDP, where each decision node has a child chance node per action. In this paper we examine environments with stochastic rewards. Each chance node represents the state and action of a MOMDP. At each chance node, the environment is sampled. For NLU-MCTS, if a new observation-reward combination is generated when sampling the environment, a new child decision node is created. This process repeats as the agent traverses the search tree. It is important to note that each chance node and its parent decision node share the same state and action. A child decision node is only created when a new observation-reward combination is received when sampling the environment. To build and traverse a search tree similar to MCTS, NLU-MCTS uses the following phases: selection, expansion, simulation and backpropagation (Section \ref{sec:MCTS}).

Now that the structure of the underlying search tree has been outlined it is possible to describe how the cumulative returns and future returns are calculated. The accrued returns is the sum of returns the NLU-MCTS algorithm receives during the execution phase from timestep 0, $t_{0}$, to timestep, $t-1$, where $\textbf{r}_t$ is the reward vector received at each timestep,
\begin{equation*}
    \textbf{R}^-_{t} = \sum_{t_{0}}^{t-1} \textbf{r}_{t}.
\end{equation*}
Given we utilise the underlying planning phases of Monte Carlo tree search, we can use the simulation phase to compute the future returns. As already mentioned during the simulation phase the agent performs a random rollout, also known as a Monte Carlo simulation, until a terminal state. NLU-MCTS utilises Monte Carlo simulations of the environment until a terminal state is reached. Therefore, the future returns can be computed from Monte Carlo simulations performed at each node during planning. Taking this into consideration the future returns, $\textbf{R}^{+}_{t}$, is the sum of the rewards received when traversing the search tree during the planning phase and Monte Carlo simulations from timestep, $t$, to a terminal node, $t_{n}$, 
\begin{equation}
    \textbf{R}^+_{t} = \sum_{t}^{t_{n}} \textbf{r}_{t}.
\end{equation}
Finally, before the utility function is applied the cumulative returns must be calculated. The cumulative returns, $\textbf{R}_t$, is the sum of the accrued returns, $\textbf{R}^-_{t}$, and the expected future returns, $\textbf{R}^+_{t}$,
\begin{equation}
    \textbf{R}_{t} = \textbf{R}^-_{t} + \textbf{R}^+_{t}.
\end{equation}
In other words, the cumulative returns is the returns received from a full policy execution. Once the cumulative returns, $\textbf{R}_{t}$, have been calculated, it is possible to compute the utility of the returns, $u(\textbf{R}_{t})$, to optimise for the ESR criterion.

As already highlighted, NLU-MCTS builds an expectimax search tree and utilises both decision and chance nodes. Over multiple iterations of the planning phase, NLU-MCTS constructs a search tree using the selection, expansion, simulation and backpropagation phases used by traditional MCTS \cite{silver2016}. We outline the NLU-MCTS algorithm in Algorithm \ref{alg:NLU-MCTS}.

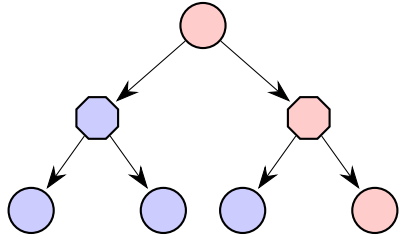
\begin{figure}
\centering
\begin{tikzpicture}[>={Stealth[width=6pt,length=9pt]}, skip/.style={draw=none}, shorten >=1pt, accepting/.style={inner sep=1pt}, auto]
  \draw (0.0pt, 0.0pt)node[circle, fill=red!20, thick, minimum height=0.6cm,minimum width=0.6cm, draw](0){};
   \draw (-40.0pt, -35.0pt)node[regular polygon,regular polygon sides=8, thick, fill=blue!20, minimum height=0.6cm,minimum width=0.6cm, draw](1){};
   \draw (40.0pt, -35.0pt)node[regular polygon,regular polygon sides=8, thick, fill=red!20, minimum height=0.6cm,minimum width=0.6cm, draw](2){};
   \draw (-15.0pt, -70.0pt)node[circle, fill=blue!20, thick, minimum height=0.6cm,minimum width=0.6cm, draw](3){};
   \draw (-65.0pt, -70.0pt)node[circle, fill=blue!20, thick, minimum height=0.6cm,minimum width=0.6cm, draw](4){};
    \draw (15.0pt, -70.0pt)node[circle, fill=blue!20, thick, minimum height=0.6cm,minimum width=0.6cm, draw](5){};
   \draw (65.0pt, -70.0pt)node[circle, fill=red!20, thick, minimum height=0.6cm,minimum width=0.6cm, draw](6){};

  \path[->] (0) edge node{} (1);
  \path[->] (1) edge node{} (3);
  \path[->] (1) edge node{} (4);
  
  \path[->] (0) edge node{} (2);
  \path[->] (2) edge node{} (5);
  \path[->] (2) edge node{} (6);


\end{tikzpicture}
\caption{During the selection phase, NLU-MCTS starts at the root node and traverses down the search tree (nodes highlighted in red). The agent traverses the search tree until a leaf decision node is found.}
\label{fig:selection_dmcts}
\end{figure}

Firstly, NLU-MCTS utilises the selection phase (Algorithm \ref{alg:NLU-MCTS_selection}, see Figure \ref{fig:selection_dmcts}), where the agent traverses the search tree starting at the current root decision node \cite{shen2019guiding}. During the selection phase, we utilise outcome selection for chance nodes and action selection for decision nodes. When the agent arrives at a chance node, we perform outcome selection where the agent simulates the environment model (Algorithm \ref{alg:NLU-MCTS_sample}). The agent then moves to the child decision node corresponding to the observation-reward combination received from the simulation \cite{shen2019guiding}. When the agent arrives at a decision node, $n_{d}$, the agent must decide which of its child chance nodes, $C_{n_d}$, to select. Therefore, NLU-MCTS selects the chance node, $n_{c}$, which maximises the UCB term:
\begin{equation}
   \text{BestChild} =  arg \ max_{n_{c} \in C_{n_d}} \text{UCB}(n_{d}, n_{c})
\label{eqn:best_child}
\end{equation}
the UCB term is defined as follows:
\begin{equation}
    \text{UCB}(n_{d}, n_{c}) = \frac{v_{n_{c}}}{N_{n_{c}}} + C \times \sqrt{\frac{ln(N_{n_{d}})}{N_{n_{c}}}},
    \label{eqn:UCB_esr}
\end{equation}
where $v_{n_{c}}$ is the total utility of the child node $n_{c}$, $N_{n_{c}}$ is the number of times child node $n_{c}$ has been visited,  $\frac{v_{n_{c}}}{N_{n_{c}}}$ is the expected utility of the child node $n_{c}$, $C$ is an exploration value, and $N_{n_{d}}$ and $N_{n_{c}}$ are the number of times $n_{d}$ and $n_{c}$ have been visited respectively. Equation \ref{eqn:UCB_esr} ensures that the agent explores areas of the tree which have not been visited often while also ensuring that the agent exploits nodes which have good returns. The agent then traverses to the chance node corresponding to the best action. The agent continues to traverse the search tree until a decision node is encountered which has not had all of is children expanded. The agent then progresses to the expansion phase (Algorithm \ref{alg:NLU-MCTS_expansion}) where the selected decision node is utilised. It is important to note that, as the agent traverses the search tree, the future returns, $\textbf{R}_{t}^{+}$, is being computed incrementally.

\begin{figure}
\centering
\begin{tikzpicture}[>={Stealth[width=6pt,length=9pt]}, skip/.style={draw=none}, shorten >=1pt, accepting/.style={inner sep=1pt}, auto]
  \draw (0.0pt, 0.0pt)node[circle, fill=blue!20, thick, minimum height=0.6cm,minimum width=0.6cm, draw](0){};
   \draw (-40.0pt, -35.0pt)node[regular polygon,regular polygon sides=8, thick, fill=blue!20, minimum height=0.6cm,minimum width=0.6cm, draw](1){};
   \draw (40.0pt, -35.0pt)node[regular polygon,regular polygon sides=8, thick, fill=blue!20, minimum height=0.6cm,minimum width=0.6cm, draw](2){};
   \draw (-15.0pt, -70.0pt)node[circle, fill=blue!20, thick, minimum height=0.6cm,minimum width=0.6cm, draw](3){};
   \draw (-65.0pt, -70.0pt)node[circle, fill=blue!20, thick, minimum height=0.6cm,minimum width=0.6cm, draw](4){};
    \draw (15.0pt, -70.0pt)node[circle, fill=blue!20, thick, minimum height=0.6cm,minimum width=0.6cm, draw](5){};
   \draw (65.0pt, -70.0pt)node[circle, fill=blue!20, thick, minimum height=0.6cm,minimum width=0.6cm, draw](6){};
   \draw (65.0pt, -105.0pt)node[regular polygon,regular polygon sides=8, fill=red!20, thick, minimum height=0.6cm,minimum width=0.6cm, draw](7){};
   \draw (65.0pt, -140.0pt)node[circle, fill=red!20, thick, minimum height=0.6cm,minimum width=0.6cm, draw](8){};

  \path[->] (0) edge node{} (1);
  \path[->] (1) edge node{} (3);
  \path[->] (1) edge node{} (4);
  
  \path[->] (0) edge node{} (2);
  \path[->] (2) edge node{} (5);
  \path[->] (2) edge node{} (6);
  \path[->] (6) edge node{} (7);
  \path[->] (7) edge node{} (8);


\end{tikzpicture}
\caption{During the expansion phase of NLU-MCTS (nodes highlighted in red), a child chance node is created. The newly generated chance node simulates the environment and creates a child decision for the corresponding reward received.}
\label{fig:expansion_dmcts}
\end{figure}
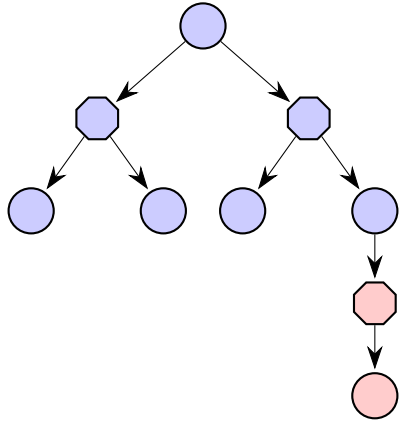

During the expansion phase (Algorithm \ref{alg:NLU-MCTS_expansion}, see Figure \ref{fig:expansion_dmcts}), the agent considers a decision node selected during the previous phase which has not had all of its children expanded. There are three steps to the expansion phase. Firstly, for the decision node, a child chance node corresponding to a previous remaining action is created for a randomly selected action. Secondly, the agent simulates the environment model for the newly created chance node. Finally, for the previously created chance node, the agent creates a child decision node corresponding to the observation-reward combination received. It is important to note that both a chance node and a decision node are generated during the expansion phase. The newly created decision node is then utilised in the next phase, known as the simulation phase.

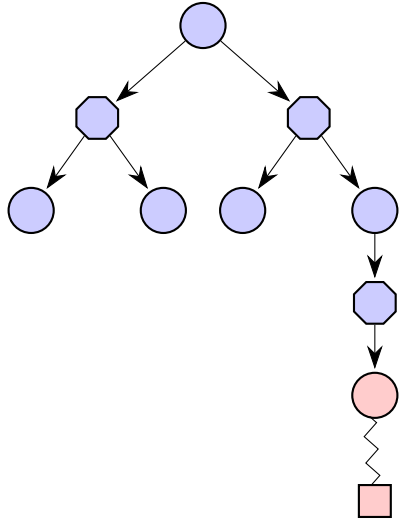
\begin{figure}
\centering
\begin{tikzpicture}[>={Stealth[width=6pt,length=9pt]}, skip/.style={draw=none}, shorten >=1pt, accepting/.style={inner sep=1pt}, auto]
  \draw (0.0pt, 0.0pt)node[circle, fill=blue!20, thick, minimum height=0.6cm,minimum width=0.6cm, draw](0){};
   \draw (-40.0pt, -35.0pt)node[regular polygon,regular polygon sides=8, thick, fill=blue!20, minimum height=0.6cm,minimum width=0.6cm, draw](1){};
   \draw (40.0pt, -35.0pt)node[regular polygon,regular polygon sides=8, thick, fill=blue!20, minimum height=0.6cm,minimum width=0.6cm, draw](2){};
   \draw (-15.0pt, -70.0pt)node[circle, fill=blue!20, thick, minimum height=0.6cm,minimum width=0.6cm, draw](3){};
   \draw (-65.0pt, -70.0pt)node[circle, fill=blue!20, thick, minimum height=0.6cm,minimum width=0.6cm, draw](4){};
    \draw (15.0pt, -70.0pt)node[circle, fill=blue!20, thick, minimum height=0.6cm,minimum width=0.6cm, draw](5){};
   \draw (65.0pt, -70.0pt)node[circle, fill=blue!20, thick, minimum height=0.6cm,minimum width=0.6cm, draw](6){};
   \draw (65.0pt, -105.0pt)node[regular polygon,regular polygon sides=8, fill=blue!20, thick, minimum height=0.6cm,minimum width=0.6cm, draw](7){};
   \draw (65.0pt, -140.0pt)node[circle, fill=red!20, thick, minimum height=0.6cm,minimum width=0.6cm, draw](8){};
   \draw[snake] (63pt, -148pt) -- (65pt, -180pt);
   \draw (65.0pt, -180.0pt)node[regular polygon,regular polygon sides=4, fill=red!20, thick, minimum height=0.6cm,minimum width=0.6cm, draw](9){};


  \path[->] (0) edge node{} (1);
  \path[->] (1) edge node{} (3);
  \path[->] (1) edge node{} (4);
  
  \path[->] (0) edge node{} (2);
  \path[->] (2) edge node{} (5);
  \path[->] (2) edge node{} (6);
  \path[->] (6) edge node{} (7);
  \path[->] (7) edge node{} (8);


\end{tikzpicture}
\caption{During the simulation phase of NLU-MCTS (nodes highlighted in red), the decision node generated in the expansion phase executes a random policy until a terminal state. Finally, the cumulative returns $\textbf{R}_{t}$ is computed.}
\label{fig:simulation_dmcts}
\end{figure}

After expansion, the created decision node must be simulated. Figure \ref{fig:simulation_dmcts} highlights the simulation phase (Algorithm \ref{alg:NLU-MCTS_simulation}) for NLU-MCTS. When a decision node is simulated, a random rollout is executed. During the rollout, a random policy is followed until a terminal state is reached. Once the simulation has completed, the cumulative returns, $\textbf{R}_{t}$, can be computed. The future returns, $\textbf{R}_{t}^{+}$, is equal to the sum of the rewards received when traversing the search tree and the returns from the random rollout in the simulation phase. The cumulative returns, $\textbf{R}_{t}$, is then computed by adding both the accrued returns, $\textbf{R}^{-}_{t}$, and the future returns, $\textbf{R}^{+}_{t}$. We note that $\textbf{R}_t$ is the same for every node during backpropagation.

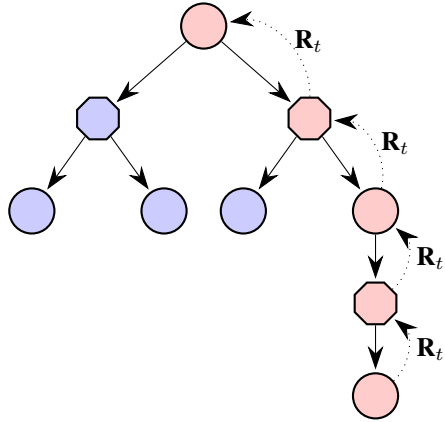
\begin{figure}
\centering
\begin{tikzpicture}[>={Stealth[width=6pt,length=9pt]}, skip/.style={draw=none}, shorten >=1pt, accepting/.style={inner sep=1pt}, auto]
  \draw (0.0pt, 0.0pt)node[circle, fill=red!20, thick, minimum height=0.6cm,minimum width=0.6cm, draw](0){};
   \draw (-40.0pt, -35.0pt)node[regular polygon,regular polygon sides=8, thick, fill=blue!20, minimum height=0.6cm,minimum width=0.6cm, draw](1){};
   \draw (40.0pt, -35.0pt)node[regular polygon,regular polygon sides=8, thick, fill=red!20, minimum height=0.6cm,minimum width=0.6cm, draw](2){};
   \draw (-15.0pt, -70.0pt)node[circle, fill=blue!20, thick, minimum height=0.6cm,minimum width=0.6cm, draw](3){};
   \draw (-65.0pt, -70.0pt)node[circle, fill=blue!20, thick, minimum height=0.6cm,minimum width=0.6cm, draw](4){};
    \draw (15.0pt, -70.0pt)node[circle, fill=blue!20, thick, minimum height=0.6cm,minimum width=0.6cm, draw](5){};
   \draw (65.0pt, -70.0pt)node[circle, fill=red!20, thick, minimum height=0.6cm,minimum width=0.6cm, draw](6){};
   \draw (65.0pt, -105.0pt)node[regular polygon,regular polygon sides=8, fill=red!20, thick, minimum height=0.6cm,minimum width=0.6cm, draw](7){};
   \draw (65.0pt, -140.0pt)node[circle, fill=red!20, thick, minimum height=0.6cm,minimum width=0.6cm, draw](8){};


  \path[->] (0) edge node{} (1);
  \path[->] (1) edge node{} (3);
  \path[->] (1) edge node{} (4);
  
  \path[->] (0) edge node{} (2);
  \path[->] (2) edge node{} (5);
  \path[->] (2) edge node{} (6);
  \path[->] (6) edge node{} (7);
  \path[->] (7) edge node{} (8);
  \path[->] (8) edge[dotted, bend right = 50] node[right]{$\textbf{R}_{t}$} (7);
  \path[->] (7) edge[dotted, bend right = 50] node[right]{$\textbf{R}_{t}$} (6);
  \path[->] (6) edge[dotted, bend right = 50] node[right]{$\textbf{R}_{t}$} (2);
  \path[->] (2) edge[dotted, bend right = 50] node[right]{$\textbf{R}_{t}$} (0);


\end{tikzpicture}
\caption{During the backpropagation phase, the cumulative returns, $\textbf{R}_{t}$,  is backpropagated to each node visited during the planning phase.}
\label{fig:backpropagation_dmcts}
\end{figure}

Figure \ref{fig:backpropagation_dmcts} and Algorithm \ref{alg:NLU-MCTS_backpropagate} outlines the backpropagation phase of NLU-MCTS. Once the simulation phase has completed, the cumulative returns, $\textbf{R}_{t}$, is backpropagated to each node visited during the previous phases of the search tree. As the agent backpropagates the cumulative returns, the agent updates the required statistic for each node.

Under the ESR criterion, the utility of the cumulative returns, u($\textbf{R}_{t}$), is computed during the backpropagation phase\footnote{To compute policies under the ESR criterion it is also possible to backpropagate the utility of the cumulative returns, u($\textbf{R}_{t}$). The relevant statistics can then be updated using the utility of the cumulative returns.} by applying the known utility function, $u$, to the cumulative returns, $\textbf{R}_{t}$. Therefore during backpropagation, the statistics at chance node are updated by updating the total utility, $v$, of the node as follows:
\begin{equation}
    v_{n_{c}} \leftarrow v_{n_{c}} + u(\textbf{R}_{t}).
\label{eqn:update_utility}
\end{equation}
The visit count for both chance node and decision nodes is also updated as follows:
\begin{equation}
    N_{n_{c}} \leftarrow N_{n_{c}} + 1,
\label{eqn:update_chance_count}
\end{equation}
\begin{equation}
    N_{n_{d}} \leftarrow N_{n_{d}} + 1.
\label{eqn:update_decision_count}
\end{equation}
The NLU-MCTS algorithm runs each step of the planning phase (selection, expansion, simulation and backpropagation) a specified number of times. We denote the number of times the planning phase is run as $n_{exec}$.
Once the NLU-MCTS algorithm has run the planning phase an $n_{exec}$ number of times, the algorithm returns the best action to take from the current root node, $n_{r}$. Under the ESR criterion, the best action, $a^{*}$, can be calculated by evaluating the expected utility of each of the current root nodes, $n_{r}$, children, $C_{n_{r}}$ and taking the action which returns the maximum expected utility as follows:
\begin{equation}
    a^{*} =  \arg \max_{n \in C_{n_{r}}} \frac{v_{n}}{N_{n}}.
    \label{eqn:nlu_mcts_bestaction}
\end{equation}

\begin{algorithm}[h]
\SetAlgoLined
$\textbf{Input}: \text{N}_{root}$ $\leftarrow$ $\text{Root node};$ $\textbf{R}_{t}^{-} \leftarrow$ $\text{Accrued returns}$\\
\textbf{Output}: Action\ $a$\ \\

\While {Not out of computation} {
$\text{N} \leftarrow N_{root}$ \\
$\textbf{R}_{t}^{+} \leftarrow$ $\text{Future returns with 0 value entry per objective}$ \\
$\text{N, } \textbf{R}_{t}^{+} \leftarrow \textbf{Selection}($N$, \textbf{R}_{t}^{+})$ \\
$\textbf{R}_{t}^{+} \leftarrow \textbf{Simulation}(\text{N}, \textbf{R}_{t}^{+})$ \\
$\textbf{R}_{t} \leftarrow \textbf{R}_{t}^{+} + \textbf{R}_{t}^{-}$ \\
$\textbf{Backpropagate}(\text{N}, \textbf{R}_{t})$
}
$\text{bestAction} \leftarrow \textbf{calculateBestAction}(\text{N}_{root}) \ \text{(Equation $\ref{eqn:nlu_mcts_bestaction}$)}$ \\
$\textbf{return} \text{ bestAction}$
 \caption{Monte Carlo Tree Search for Nonlinear Utility Functions}
 \label{alg:NLU-MCTS}
\end{algorithm}


\begin{algorithm}[h]
\SetAlgoLined
\textbf{Input}: \text{N}\ $\leftarrow$ Node\ in\ the\ tree;\ $\textbf{R}_{t}^{+} \leftarrow$ $\text{Future returns}$\\ 
\If{\text{N}\ is\ terminal}{\textbf{return}\ \text{N}, $\textbf{R}_{t}^{+}$}
\If{\text{\text{N} is a chance node}}{
$\text{N}, \textbf{R}_{t}^{+} \leftarrow \textbf{Sample}(\text{N}, \textbf{R}_{t}^{+})$
}
\If{$\text{N has children to expand}$}
{
$\text{N}^{*},\ \textbf{R}_{t}^{+} \leftarrow \textbf{Expansion}(\text{N}, \textbf{R}_{t}^{+})$ \\
$\textbf{return} \ \text{N}^{*}, \textbf{R}_{t}^{+}$
}
$\text{N}\leftarrow \textbf{BestChild} \ \text{(Equation \ \ref{eqn:best_child})}$\\
$\textbf{Selection}(\text{N}, \textbf{R}_{t}^{+})$
\caption{Selection}
 \label{alg:NLU-MCTS_selection}
\end{algorithm}


\begin{algorithm}[h]
\SetAlgoLined
$\textbf{Input}: \text{N} \leftarrow \text{Node in the tree;} \ \textbf{R}_{t}^{+} \leftarrow \text{Future returns}$\\ 

$\text{N}^{*} \leftarrow \text{a\ new\ child\ chance\ node\ for\ a\ remaining\ action}$ \\
$\text{add } \text{N}^{*} \text{ to N's children}$ \\
$\text{N}^{*}\text{,} \ \textbf{R}_{t}^{+} \leftarrow \textbf{Sample}(\text{N}^{*}, \textbf{R}_{t}^{+})$ \\
$\textbf{return} \ \text{N}^{*},$ $\textbf{R}_{t}^{+}$

\caption{Expansion}
 \label{alg:NLU-MCTS_expansion}
\end{algorithm}


\begin{algorithm}[H]
\SetAlgoLined
\textbf{Input}: $\text{N}\ \leftarrow$ $\text{Chance Node; }$ $\textbf{R}_{t}^{+}$ $\leftarrow$ $\text{Future returns}$ \\

$\text{observation, reward}$ $\leftarrow$ simulate\ agent\ environment\ for\ $\text{N}$ \\
$\textbf{R}_{t}^{+}$ $\leftarrow \textbf{R}_{t}^{+} + \text{reward} $ \\
\For{$DN$\ in $N$.Children}{
\If{(DN.observation, DN.reward) = (observation, reward)}
{
$\textbf{return}\ \text{DN, } \textbf{R}_{t}^{+}$
}
}
DN\ $\leftarrow$ N\ create\ child\ decision\ node\ (observation, reward) \\
$\textbf{return}\ \text{DN, } \textbf{R}_{t}^{+}$

 \caption{Sample}
 \label{alg:NLU-MCTS_sample}
\end{algorithm}


\begin{algorithm}[H]
\SetAlgoLined
$\textbf{Input}: \text{N } \leftarrow \text{Node in the tree; } \text{s } \leftarrow \text{Node.state; } \textbf{R}_{t}^{+} \leftarrow \text{Future returns}$\\
\While{$\text{s is not terminal}$}
{
$\text{a } \leftarrow \text{random action}$\\
$\text{s, }\textbf{r}_{t} \leftarrow \text{env(s, a)}$\\
$\textbf{R}_{t}^{+} \leftarrow \textbf{R}_{t}^{+}\ +\ \mathbf{r}_{t}$ \\
}
$\textbf{return} \text{ } \textbf{R}_{t}^{+}$

\caption{Simulation}
 \label{alg:NLU-MCTS_simulation}
\end{algorithm}

\begin{algorithm}[h]
\SetAlgoLined
$\textbf{Input}: \text{N} \leftarrow \text{Node in the tree; }$
$\textbf{R}_{t} \leftarrow \text{Cumulative returns}$\\

\While{$\text{N is not null}$}
{
$\textbf{UpdateStatistics}\text{(N, } \textbf{R}_{t})$ \text{(Equations \ref{eqn:update_utility}, \ref{eqn:update_chance_count} and \ref{eqn:update_decision_count})} \\
$\text{N} \leftarrow \text{N.parent}$ \\

}
\caption{Backpropagate}
 \label{alg:NLU-MCTS_backpropagate}
\end{algorithm}

\section{Distributional Monte Carlo Tree Search}
\label{sec:DMCTS}
Monte Carlo tree search for nonlinear utility functions (NLU-MCTS) utilises the UCB statistic to explore during planning. However, Thompson sampling methods have been shown to outperform UCB methods in bandit settings \cite{russo2014learning,chapelle2011thompson}. Therefore, to exploit the potential performance increases associated with Thompson sampling methods, we present a novel distributional Monte Carlo tree search algorithm (DMCTS) that learns a posterior distributions over the expected utility of the returns.

\begin{algorithm}[h]
\SetAlgoLined
$\textbf{Input}: N_{root}$ $\leftarrow$ $\text{Root node};$ $\textbf{R}_{t}^{-} \leftarrow$ $\text{Accrued returns}$\\
\textbf{Output}: Action\ $a$\ \\
\While {Not out of computation} {
$\text{N} \leftarrow \text{N}_{root}$ \\
$\textbf{R}_{t}^{+} \leftarrow$ $\text{Future returns with 0 value entry per objective}$ \\
$\text{N, } \textbf{R}_{t}^{+} \leftarrow \textbf{Selection}($N$, \textbf{R}_{t}^{+})$ \\
$\textbf{R}_{t}^{+} \leftarrow \textbf{Simulation}(\text{N}, \textbf{R}_{t}^{+})$ \\
$\textbf{R}_{t} \leftarrow \textbf{R}_{t}^{+} + \textbf{R}_{t}^{-}$ \\
$\textbf{Backpropagate}(\text{N}, \textbf{R}_{t})$
}
$\text{bestAction} \leftarrow \textbf{calculateBestAction}(\text{N}_{root})$\label{dmcts_best_action} \\
$\textbf{return} \text{ bestAction}$
 \caption{Distributional Monte Carlo Tree Search}
 \label{alg:DMCTS}
\end{algorithm}

\begin{algorithm}
\SetAlgoLined
\textbf{Input}: N\ $\leftarrow$ Node\ in\ the\ tree;\ $\textbf{R}_{t}^{+} \leftarrow$ $\text{Future returns}$\\ 
\If{N\ is\ terminal}{\textbf{return}\ N, $\textbf{R}_{t}^{+}$}
\If{\text{N is a chance node}}{
$\text{N}, \textbf{R}_{t}^{+} \leftarrow \textbf{Sample}(\text{N}, \textbf{R}_{t}^{+})$
}
\If{$\text{N has children to expand}$}
{
$\text{N}^{*},\ \textbf{R}_{t}^{+} \leftarrow \textbf{Expansion}(\text{N}, \textbf{R}_{t}^{+})$ \\
$\textbf{return} \ \text{N}^{*}, \textbf{R}_{t}^{+}$
}
$\text{N} \leftarrow \textbf{ThompsonSampling}$\\
$\textbf{Selection}(\text{N}, \textbf{R}_{t}^{+})$
\caption{Selection}
 \label{alg:DMCTS_selection}
\end{algorithm}

Firstly, it is important to discuss how DMCTS (Algorithm \ref{alg:DMCTS}) builds an underlying search tree. DMCTS builds an expectimax search tree using the same planning phase as NLU-MCTS (see Section \ref{sec:nonlinear_MCTS}). However, DMCTS takes a distributional approach to decision making.

DMCTS aims to maintain a posterior distribution over the expected utility of the returns at each chance node.  However, because the utility function may be nonlinear, a parametric form of the posterior distribution may not exist. Since a bootstrap distribution can be used to approximate a posterior \cite{efron2012,newton1995}, it is much more suitable to maintain a bootstrap distribution over the expected utility of the returns at each chance node. 

\begin{figure}

\centering
\begin{tikzpicture}[line cap=round, line join=round]
\draw (132.0pt,184.0pt) node[below](0) {} ;
\draw (200.0pt,185.0pt) node[below](1) {\fontsize{30}{30}\selectfont$\frac{\alpha_{j}}{\beta_{j}}$} ;
\path[->] (1) edge[bend right = 50] node[right]{} (0);
\begin{axis}[
xlabel={\textbf{Bootstrap replicates (j)}},
ylabel={\textbf{Expected utility}},
ytick={0, 0.1, 0.2},
xtick={0,1,2,3,4,5,6,7,8,9,10,11,12,13,14,15}, 
xticklabels={$j_{0}$, $j_{1}$, $j_{2}$, $j_{3}$, $j_{4}$, $j_{5}$, $j_{6}$, $j_{7}$, $j_{8}$, $j_{9}$, 
$j_{10}$, $j_{11}$, $j_{12}$, $j_{13}$, $j_{14}$, $j_{15}$}
]
\addplot [ybar, domain=0:7, samples=8, fill=blue!50!cyan, draw=none] 
  (x, {norm(x, 3.5, 0.25)});

\end{axis}
\end{tikzpicture}
\caption{A bootstrap distribution learned by DMCTS with the number of bootstrap replicates, $J$, set to $8$. The expected utility for each bootstrap replicate, $j$, can be calculated by $\frac{\alpha_{j}}{\beta_{j}}$. For example, the expected utility for bootstrap replicate $j_{4}$ can be calculated as follows: $\mathbb{E}(u(j_{4})) = \frac{\alpha_{j_{4}}}{\beta_{j_{4}}}$.}
\label{fig:bts}
\end{figure}
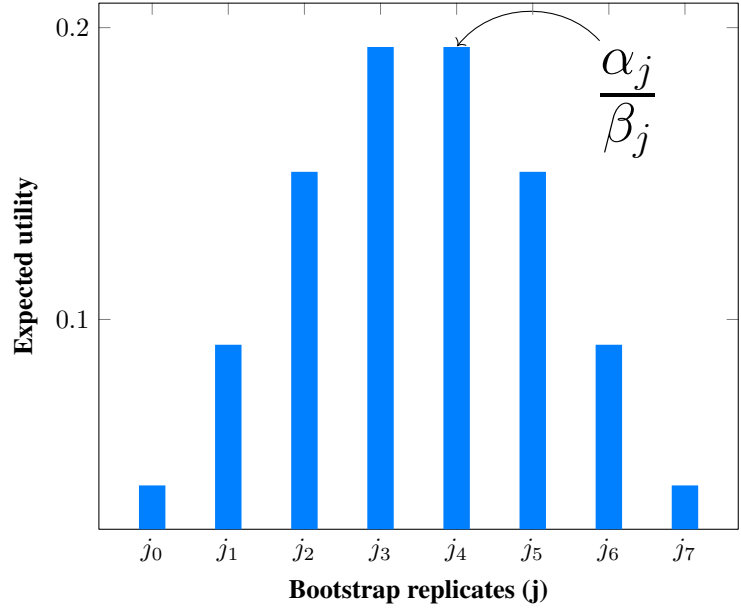

Each bootstrap distribution contains a number of bootstrap replicates, $j \in  \{1, ..., J\}$ \cite{EcklesKaptein2014} (see Section \ref{sec:BTS}). It is important to note the number of bootstrap replicates, $J$, is a hyperparameter that can be tuned for exploration \cite{EcklesKaptein2014}. Each bootstrap replicate, $j$, in the bootstrap distribution has two parameters, $\alpha_{j}$\footnote{In this work our use of $\alpha$ differs slightly from that of Kaptein and Eckles \cite{EcklesKaptein2014}. We utilise $\alpha$ to track the sum of the utility, which can then be utilised to compute the expectation. Whereas, Kaptein and Eckles utilise $\alpha$ as a count for the returns of a Bernoulli bandit.} and $\beta_{j}$, where $\frac{\alpha_{j}}{\beta_{j}}$ is the expected utility for replicate $j$. On initialisation of a new node, for each bootstrap replicate, $j$, the parameters $\alpha_{j}$ and $\beta_{j}$ are both set to $1$. Moreover, $\alpha_{j}$ can be set to positive or negative values to increase initial exploration without a computational cost. Figure \ref{fig:bts} outlines a bootstrap distribution learned by the DMCTS algorithm. For ESR settings, the expected utility of each bootstrap replicate, $j$, can be computed as follows:
\begin{equation}
    \mathbb{E}(u(j)) = \frac{\alpha_{j}}{\beta_{j}}.
\end{equation}
It is important to note that, similarly to NLU-MCTS, DMCTS requires the utility function of the user to be known a priori. The bootstrap distribution is updated during the backpropagation phase of the DMCTS algorithm.

\begin{algorithm}
\SetAlgoLined
$\textbf{Input}: \text{N} \leftarrow \text{Node in the tree}$ \\
$\textbf{Input}: \textbf{R}_{t}\ \leftarrow \text{Cumulative returns}$ \\

\While{$\text{N is not null}$}
{
$\textbf{UpdateDistribution}(\text{N}, \textbf{R}_{t})$ \\
$\text{N} \leftarrow \text{N.parent}$ \\

}
\caption{Backpropagate}
 \label{alg:DMCTS_backpropagate}
\end{algorithm}



During the backpropagation phase (Algorithm \ref{alg:DMCTS_backpropagate}) the cumulative returns is backpropagated and the bootstrap distribution at each chance node is updated. Algorithm \ref{alg:DMCTS_UpdateDistribution_ESR} outlines how a bootstrap distribution for a node is updated for the ESR criterion. In this paper, we do not use discounting as we perform evaluations only on finite horizon tasks. We note that DMCTS can easily be adapted to discounted settings. At chance node, $i$, for each bootstrap replicate, $j$, a coin flip is simulated (See Algorithm \ref{alg:DMCTS_UpdateDistribution_ESR}, Line \ref{alg:DMCTS_UpdateDistribution_ESR:line:bernoulli}). If the result of the coin flip is equal to $1$ (heads), $\alpha_{ij}$ and $\beta_{ij}$ are updated:
\begin{equation*}
    \alpha_{ij} \leftarrow \alpha_{ij} +  u(\textbf{R}_t)
\end{equation*}
\begin{equation*}
    \beta_{ij} \leftarrow \beta_{ij} + 1
\end{equation*}

To select actions while planning (Algorithm \ref{alg:DMCTS_selection}), we use the previously computed statistics. At each timestep the agent must choose which action to execute in order to traverse the search tree (as outlined in Algorithm \ref{alg:DMCTS_ThompsonSample_ESR}). At decision node $n$, we select an action by sampling the bootstrap distribution at each child chance node, $i$. For each sampled bootstrap replicate, $j$, the $\alpha_{ij}$ and $\beta_{ij}$ values are retrieved and $\frac{\alpha_{ij}}{\beta_{ij}}$ is computed. 
Since the following approximation is true,
\begin{equation}
    \frac{\alpha_{ij}}{\beta_{ij}} \equiv \mathbb{E}[u(\textbf{R}^-_{t} + \textbf{R}^+_{t})],
\label{eqn:alphabeta_ESR}
\end{equation}
by maximising over $i$ in Equation \ref{eqn:alphabeta_ESR}, we select an action corresponding to $j$ approximately proportional to the probability of that action being optimal -- per the Bootstrap Thompson Sampling exploration strategy. The agent then executes the action, $a^*$, which corresponds to the following:
\begin{equation}
 a^* = \arg \max_{i} \frac{\alpha_{ij}}{\beta_{ij}}.
 \label{eqn:dmcts_best_action}
\end{equation}
We note that, at execution time, we can calculate the best action (Algorithm \ref{alg:DMCTS}, Line \ref{dmcts_best_action}) by simply selecting the overall maximising action by averaging over all the acquired data, thereby maximising the ESR criterion:
\begin{equation}
    ESR = \mathbb{E}[u(\textbf{R}^-_{t} + \textbf{R}^+_{t})].
\label{eqn:ESR_DMCTS}
\end{equation}

\begin{algorithm}
\SetAlgoLined
\textbf{Input}: i\ $\leftarrow$ Node\ in\ the\ tree;\ 
$\textbf{R}_{t}\ \leftarrow$ Cumulative\ Returns\ \\
J\ $\leftarrow$ node.bootstrapDistribution \\
\For{j,\ ...,\ $J$\ bootstrap\ replicates}{
Sample\ $d_{j}$\ from\ Bernoulli($\frac{1}{2}$) \label{alg:DMCTS_UpdateDistribution_ESR:line:bernoulli} \\
\If{$d_{j}$ = 1}{
$\alpha_{ij} \leftarrow \alpha_{ij} + u(\textbf{R}_{t})$ \\
$\beta_{ij} \leftarrow \beta_{ij} + 1$ \\
}}
\caption{UpdateDistribution}
\label{alg:DMCTS_UpdateDistribution_ESR}
\end{algorithm}
\begin{algorithm}
\SetAlgoLined
\textbf{Input}: n\ $\leftarrow$ Node\ in\ the\ tree\ \\
\textbf{Require}: $\alpha$,\ $\beta$\ prior\ parameters\ \\
$\alpha_{ij}$ := $\alpha$,\ $\beta_{ij}$ := $\beta$\  \{For\ each\ n\ child,\ $i$,\ and\ each\ bootstrap\ replicate,\ $j$ \} \\
\For{$i$\ in\ n.children} 
{
Sample\ $j$\ from\ uniform\ 1,\ ...,\ $J$\ bootstrap\ replicates\\
Retrieve\ $\alpha_{ij},\ \beta_{ij}$ \\
}
maxChild\ = $\arg \max_{i} \frac{\alpha_{ij}}{\beta_{ij}} $\\
\textbf{return} maxChild\ or\ maxChild.action
\caption{ThompsonSample}
\label{alg:DMCTS_ThompsonSample_ESR}
\end{algorithm}

Using the outlined algorithm, DMCTS is able to learn policies for risk-aware settings and under ESR for multi-objective settings. In Section \ref{sec:Experiments}, we evaluate DMCTS for risk-aware settings and multi-objective settings for the ESR criterion.

\section{Experiments}
In order to evaluate NLU-MCTS and DMCTS, we test both algorithms in multiple settings. Firstly, we perform an ablative study to outline the effect on computation and performance the $J$ parameter has when computing the BTS distribution for DMCTS. We then evaluate NLU-MCTS and DMCTS in a risk-aware setting. Finally, we evaluate both algorithms in multi-objective settings under the ESR criterion. In multi-objective settings, we test our algorithms on variants of standard benchmark problems from the MORL literature.

We also evaluate NLU-MCTS and DMCTS against two other state-of-the-art RL algorithms: Expected Utility Policy Gradient (EUPG) \cite{roijers2018multi} and C51 \cite{bellemare2017distributional}. EUPG is the only MORL algorithm that can compute policies under the ESR criterion and is therefore the state-of-the-art performance in this setting \cite{roijers2018multi}. We use C51 as a baseline algorithm for our evaluation of DMCTS given C51 is a distributional RL algorithm and has achieved state-of-the-art performance \cite{bellemare2017distributional}. 

At each timestep for NLU-MCTS and DMCTS, the planning phase is performed multiple times before an action is selected during the execution phase. To fairly evaluate all other algorithms against NLU-MCTS and DMCTS, we have altered each benchmark algorithm to have the same number of policy executions of each environment at each timestep as NLU-MCTS and DMCTS. At each timestep, each algorithm gets $n_{exec}$ full policy executions worth of learning from that state and timestep onward. Therefore, if $n_{exec} = 10$, NLU-MCTS and DMCTS perform the planning phase ten times before selecting an action. To ensure a C51 and EUPG get the same opportunity to learn, both algorithms are altered to execute a policy $n_{exec}$ number of times from the current state. For the other algorithms (except NLU-MCTS and DMCTS) this has the effect of increasing the learning speed. The number of policy executions $n_{exec}$ varies for each problem domain. All experiments are averaged over 10 runs.
\label{sec:Experiments}

\subsection{Ablation Study}
Before we evaluate both NLU-MCTS and DMCTS in risk-aware and multi-objective sequential decision making problems, we empirically evaluate how the Bootstrap Thompson Sampling (BTS) parameter settings affect performance and run time. We also provide a visualisation that shows how a BTS distribution is updated over time to estimate the underlying posterior distribution over the expected utility. Finally, we evaluate the performance of DMCTS under different J values in a MOMDP, to highlight how the selection of the J value can effect performance in sequential settings.

\subsubsection{Bootstrap Thompson Sampling J Values \& Runtime}
To illustrate how a BTS distribution evolves over time, we update a single BTS distribution based on the returns of a simple multi-objective bandit. In this setting the bandit has one arm, where there is a $0.5$ chance of receiving the following return:  $\textbf{r} = [1, 1]$, and a $0.5$ chance of receiving the following return:  $\textbf{r} = [0, 0]$. The returns are then scalarised using the following utility function:
\begin{equation}
    u = r_{1} r_{2},
\end{equation}
where $r_{1}$ and $r_{2}$ are the returns for objective $1$ and objective $2$ respectively. In this example, expected utility is $0.5$.

\begin{figure}
\includegraphics[width=0.275\textwidth]{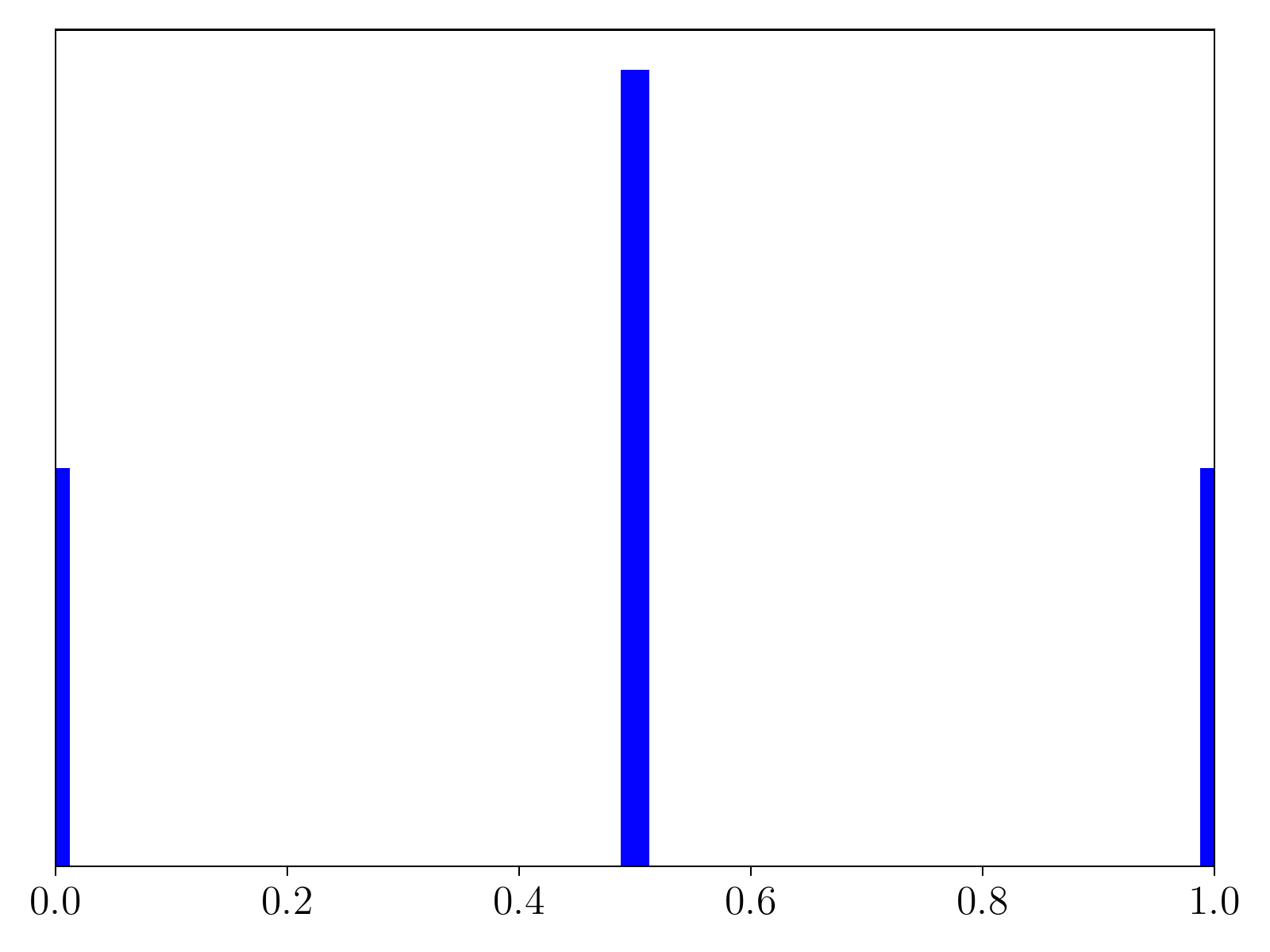}
\hspace{\fill}
\includegraphics[width=0.275\textwidth]{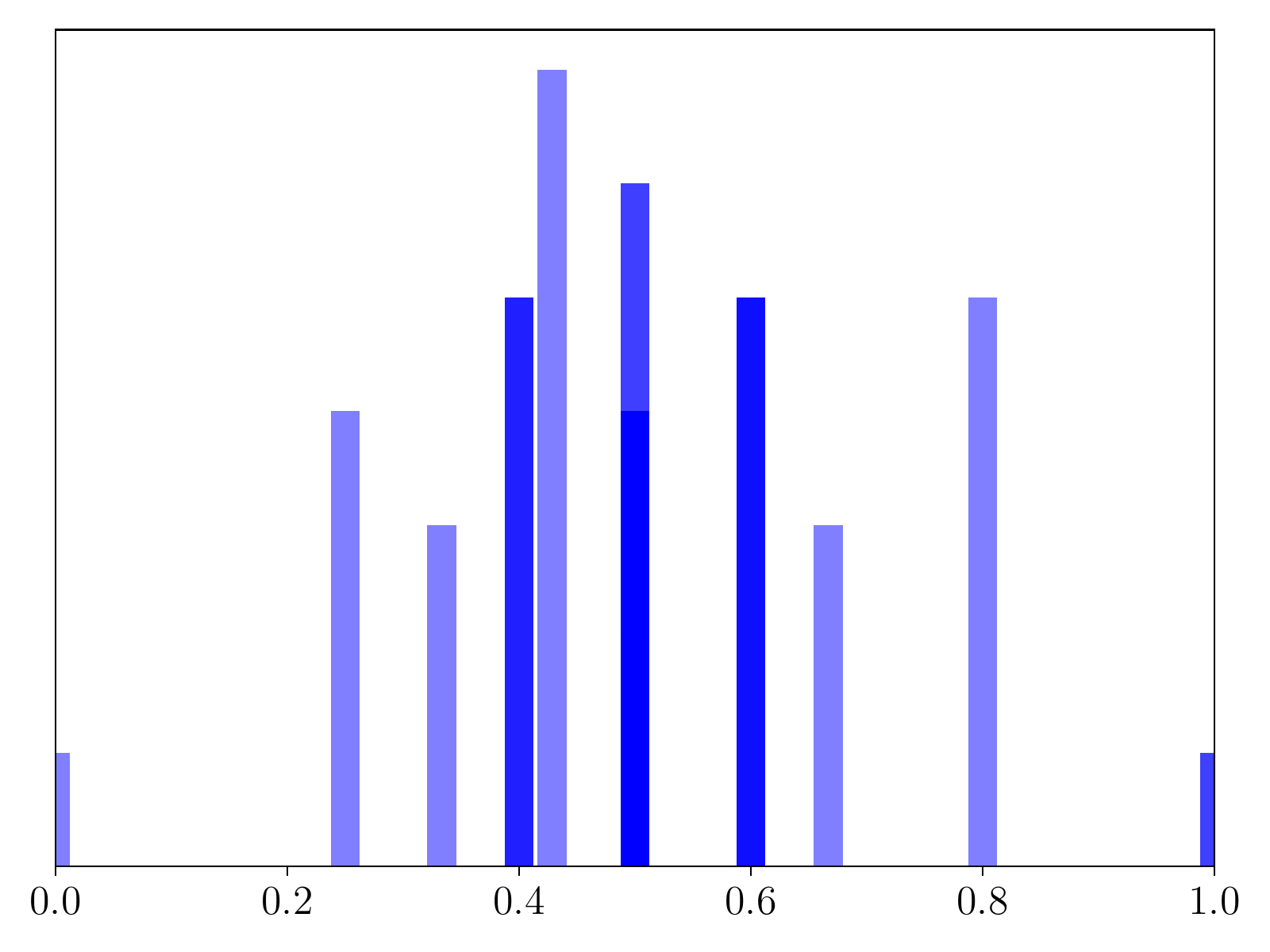}
\hspace{\fill}
\includegraphics[width=0.275\textwidth]{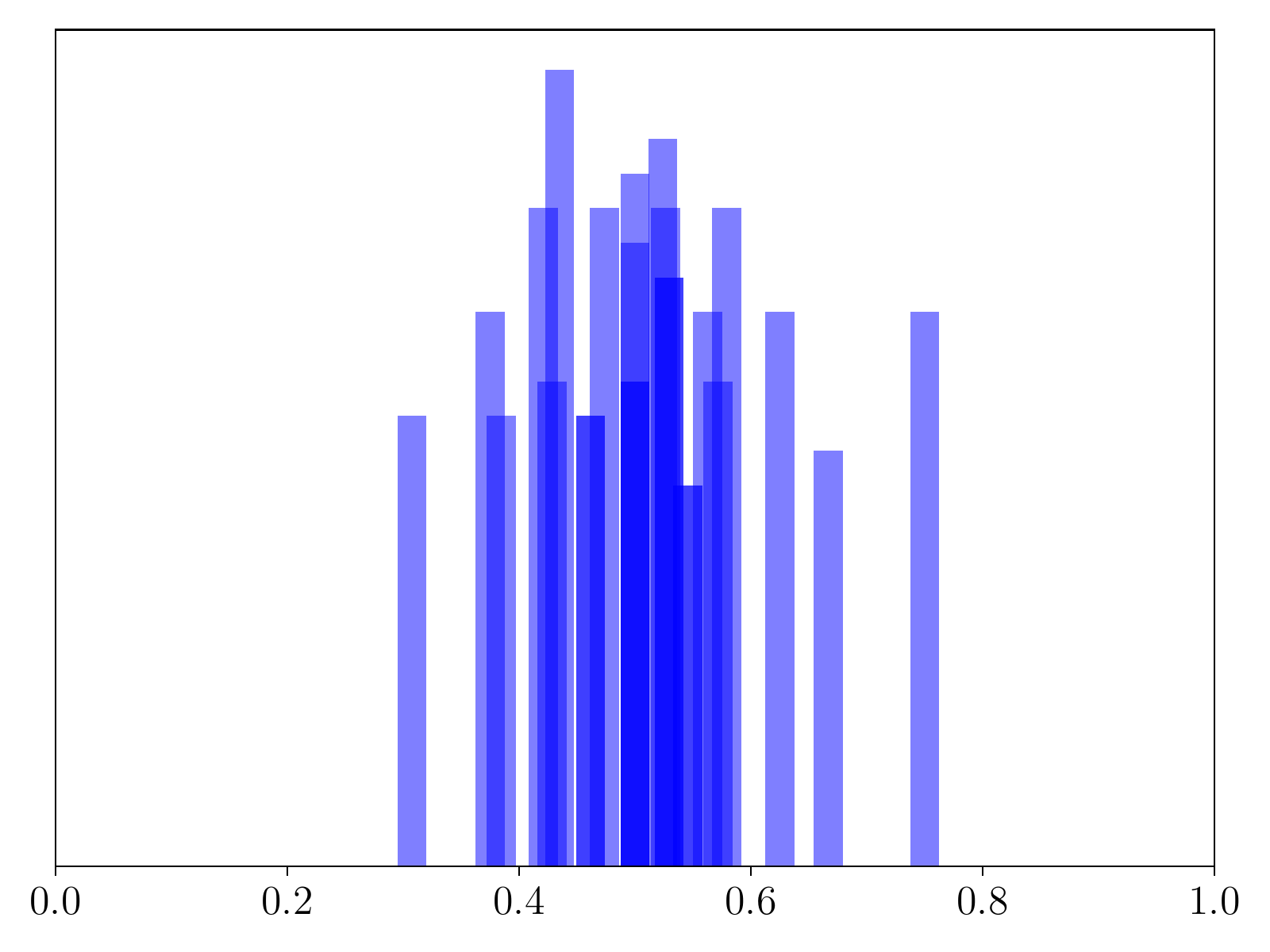}
\hspace{\fill}
\includegraphics[width=0.275\textwidth]{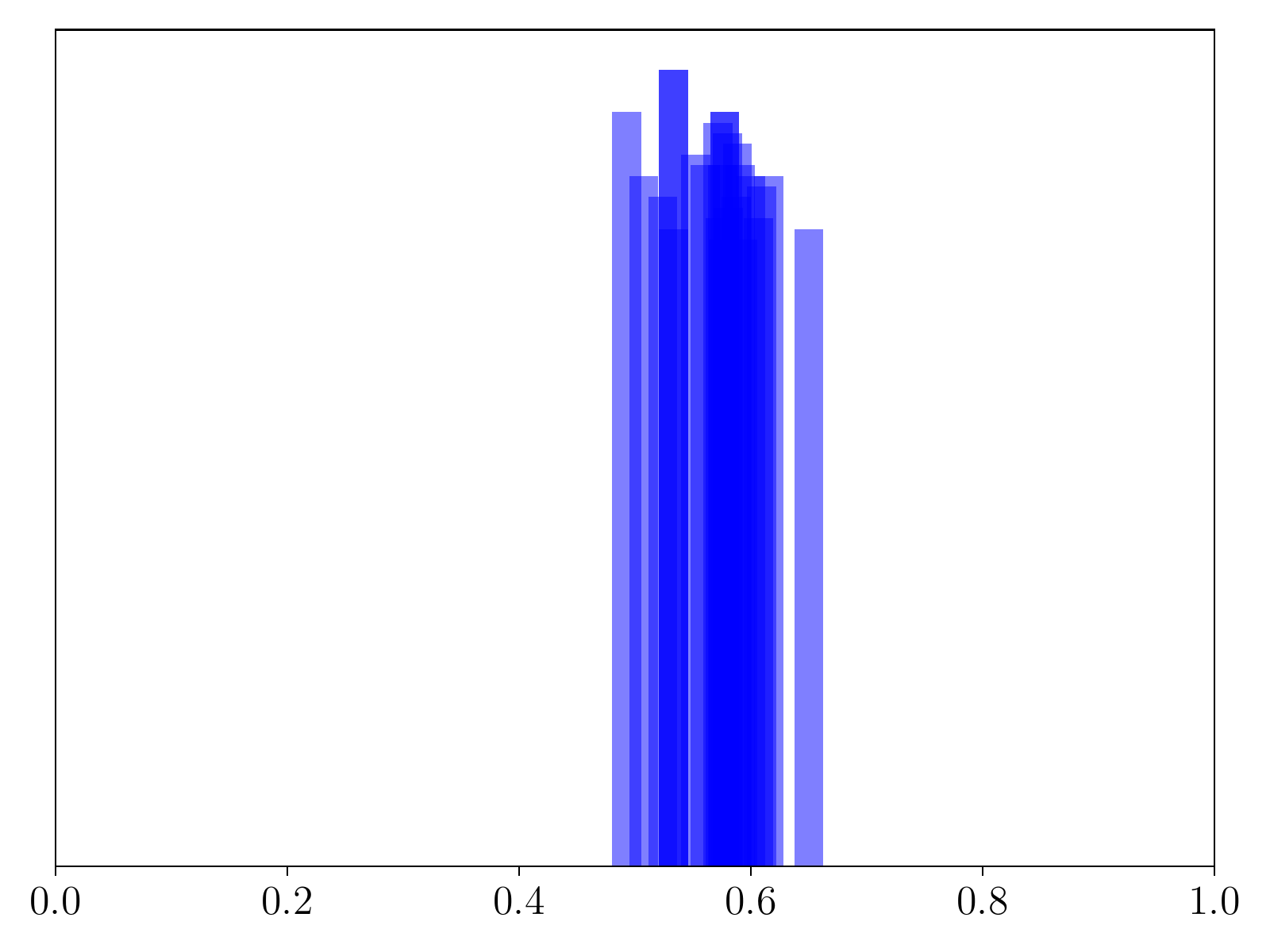}
\hspace{\fill}
\includegraphics[width=0.275\textwidth]{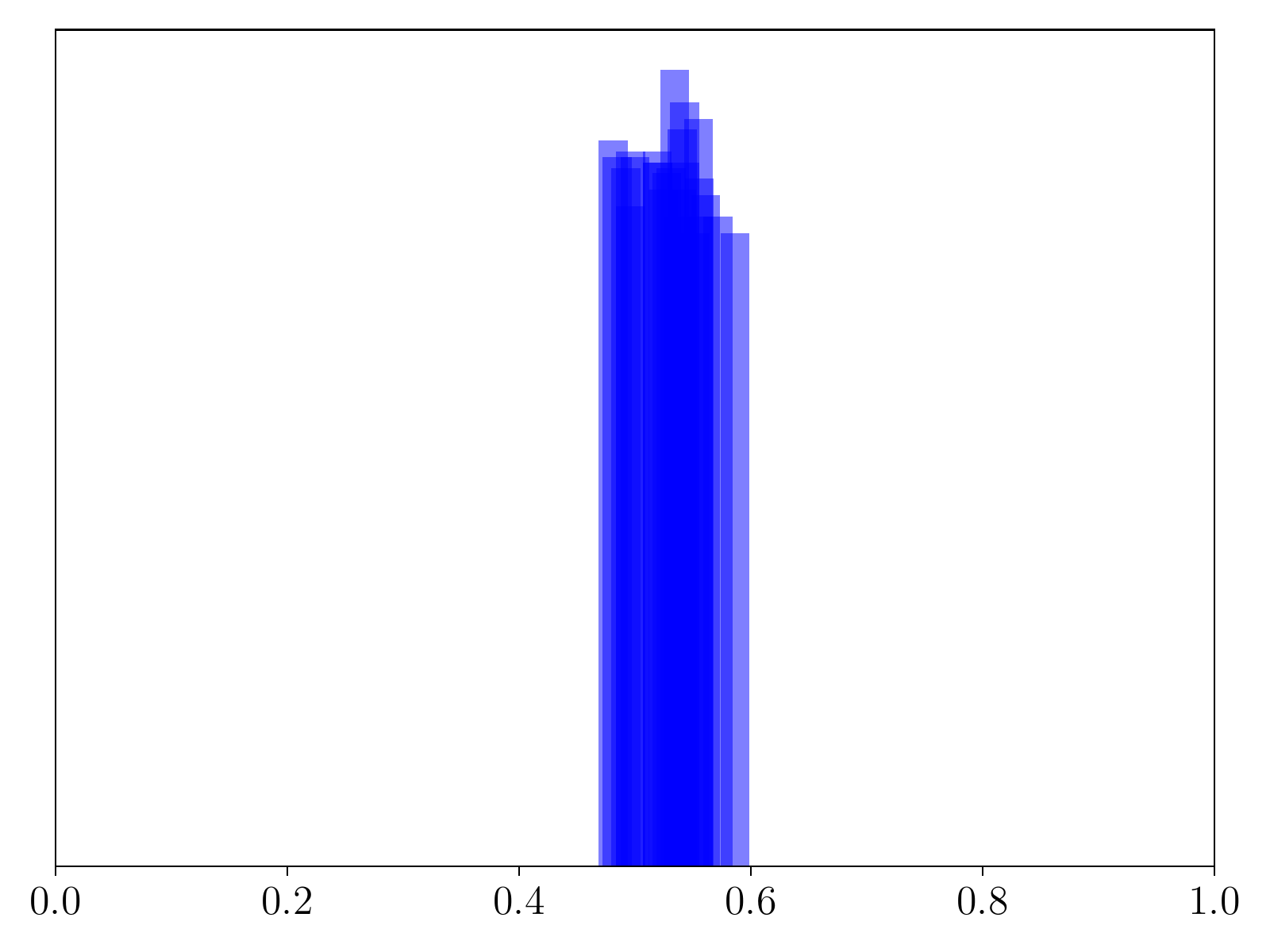}
\hspace{\fill}
\includegraphics[width=0.275\textwidth]{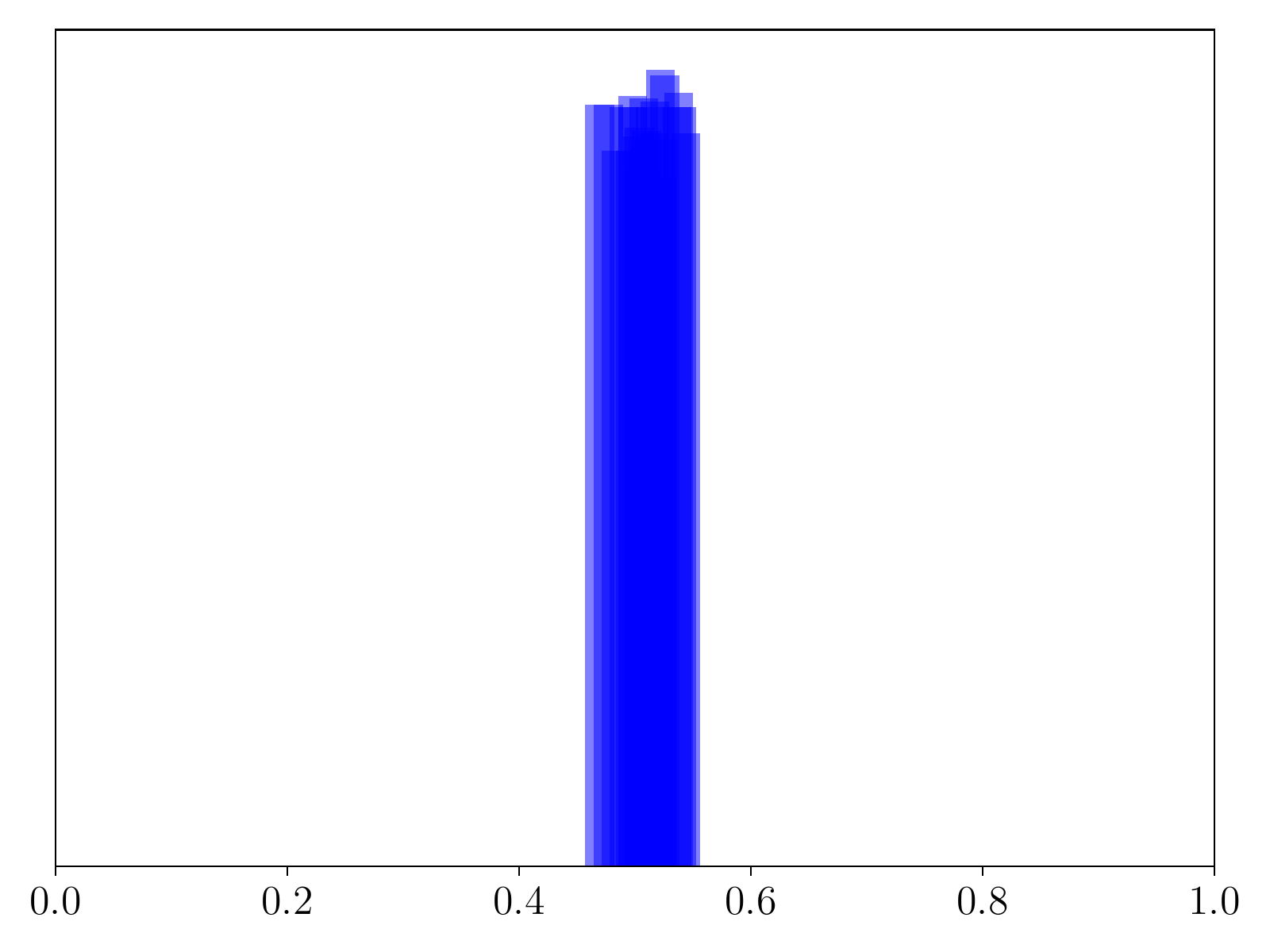}

\caption{A BTS distribution after 1, 8, 32, 128, 250 \& 300 updates. After 500 updates the distribution converges to the correct expected utility, where expected utility is on the x-axis.}\label{fig:xyz}
\label{fig:bts_updates}
\end{figure}

Using this bandit we update a single BTS distribution and show how the distribution evolves over a number of updates using the utility of the returns. Figure \ref{fig:bts_updates} outlines how a BTS distribution with $25$ bootstrap replicates evolves after $1, 8, 32, 128, 250 \ \& \ 500$ updates.

\begin{figure}
    \centering
    \includegraphics[width=0.65\textwidth]{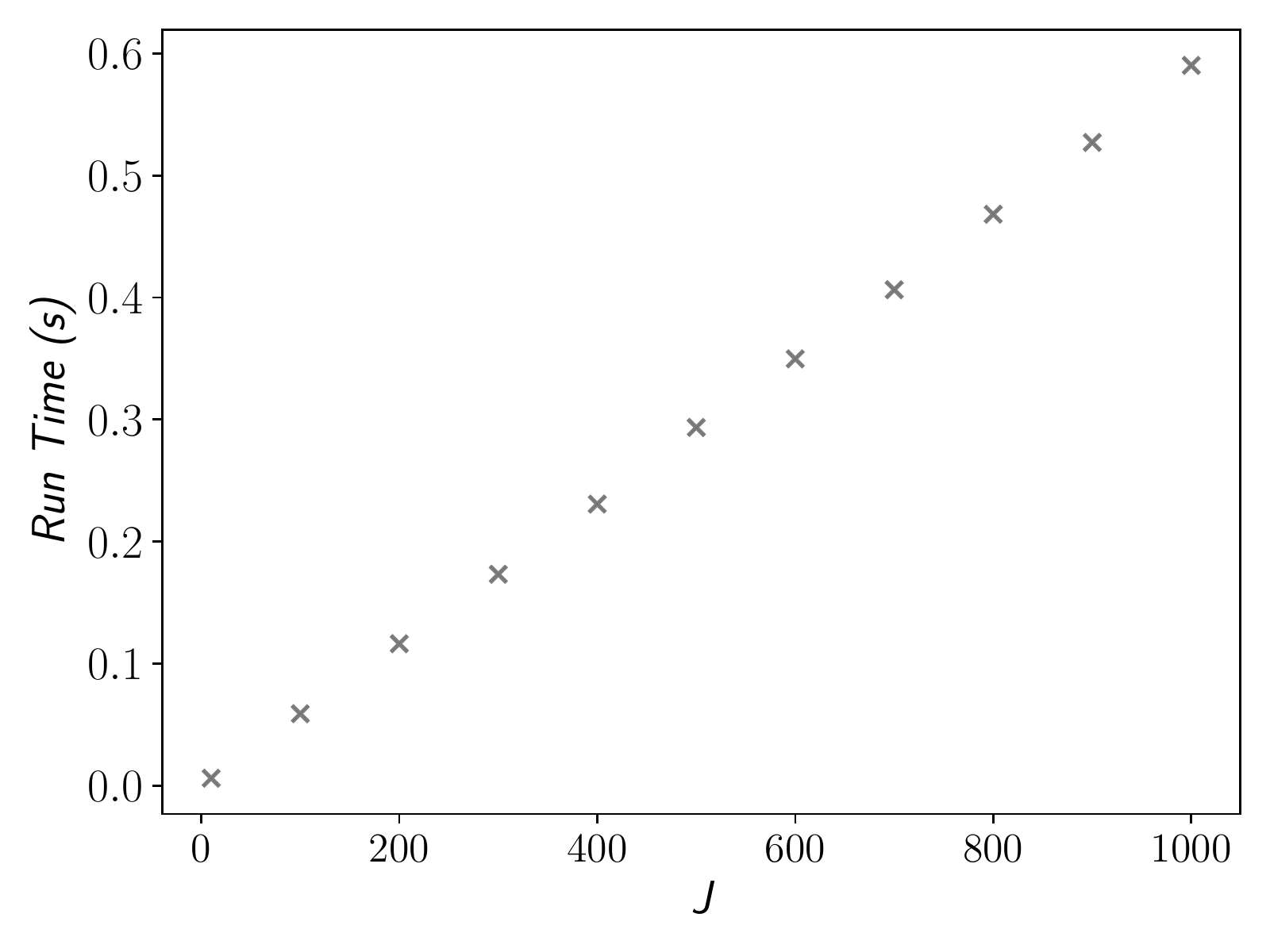}
    \caption{The run time is seconds required to complete $1,000$ updates of a BTS distribution for different $J$ values. The run time required increases linearly with the increase in the $J$ value.}
    \label{fig:bandit_runtime}
\end{figure}

Next, we investigate the computational run time for a BTS distribution with varying number of replicates, $J$. To evaluate the run time for each chosen $J$ value we compute the time in seconds taken to perform $1,000$ updates of a BTS distribution. This experiment was performed $10$ times for each $J$ value and the average run time was computed. To evaluate the run time we use the following $J$ values: $10, 100, 200, 300, 400, 500, 600, 700, 800, 900 \ \& \ 1,000$ and present the results in Figure \ref{fig:bandit_runtime}.
Figure \ref{fig:bandit_runtime} shows that the run time in seconds increases linearly with the number of replicates $J$. Therefore, the hyperparameter $J$ can have an impact on the run time of the algorithm and therefore should be taken into consideration in order to optimise performance. Next we will evaluate the performance of a BTS distribution for multiple $J$ values in a multi-objective multi-armed bandit setting. By comparing run time and performance it should be possible to determine which $J$ values can be selected for good performance and efficiency.

\subsubsection{Bootstrap Thompson Sampling J Values \& Performance}
To investigate the effect the hyperparameter $J$ on the performance of DMCTS we consider a multi-objective multi-armed bandit (MOMAB) setting where a BTS distribution is utilised per arm to determine which arm is optimal for a given utility function.

We utilise a MOMAB setting from the literature \cite{roijers2021momab}, with 5 arms, and each of the following ground truth mean vectors: (0, 0.8), (0.4, 0.4), (0.8, 0.0) and (0.9, 0.1). Each reward distribution is multi-variate Gaussian with correlations 0 and in-objective variance 0.0005 \cite{roijers2021momab}. We utilise the following utility function:
\begin{equation}
    u = 6.25 \ max(r_{0}, 0) \ max(r_{1}, 0).
\end{equation}
In this setting, the arm with mean vector (0.4, 0.4) is optimal and returns an expected utility of $1$. We run BTS for $10,000$ trials for the following $J$ values: $10, 100, 500 \ \& \ 1,000$.

\begin{figure}[h]
    \centering
    \includegraphics[width=0.65\textwidth]{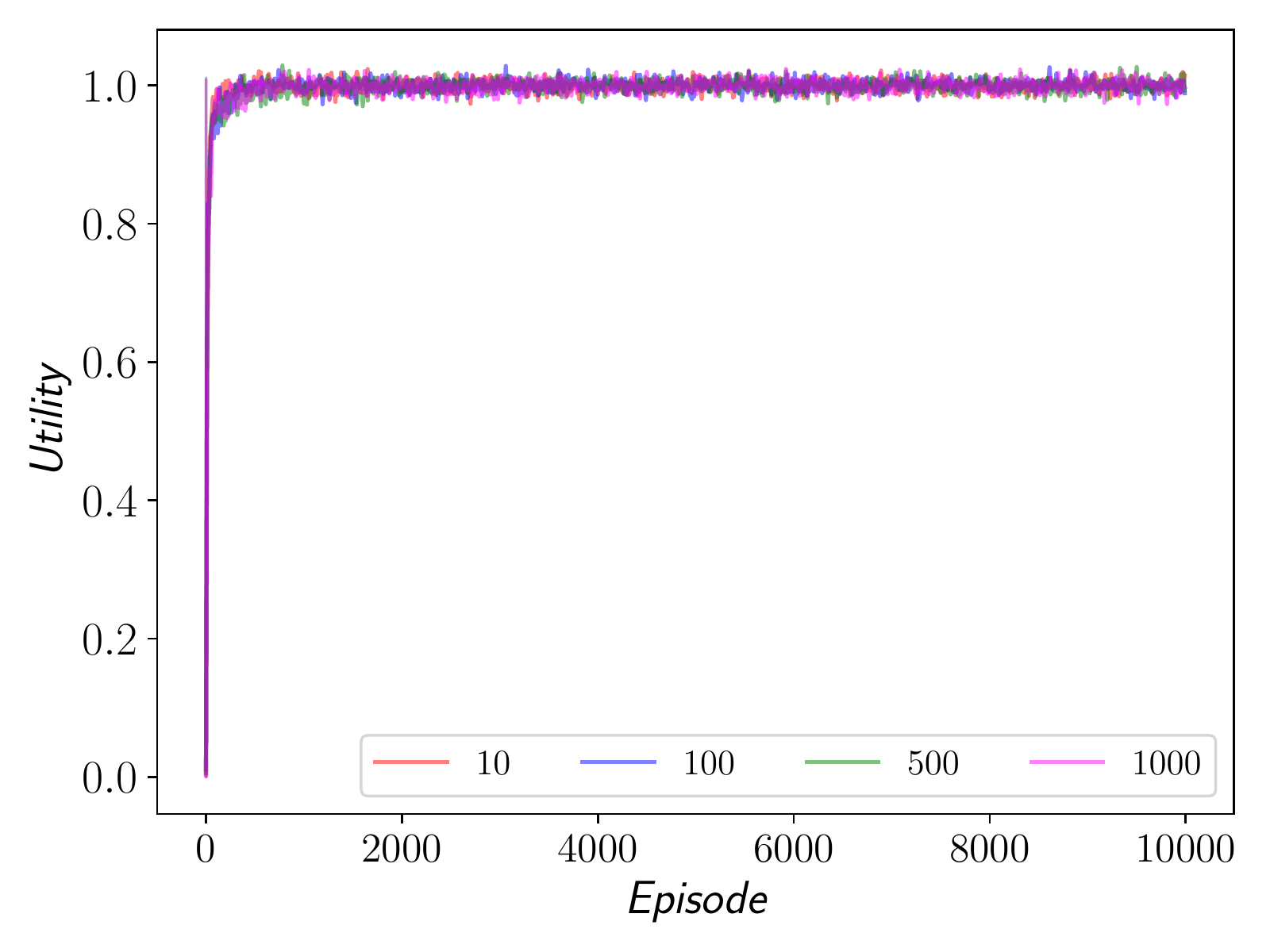}
    \caption{The performance of BTS algorithm is a multi-objective multi-armed bandit setting for different $J$ values. For each $J$ value the algorithm converges to the optimal utility of $1$.}
    \label{fig:momab_utility}
\end{figure}

Figure \ref{fig:momab_utility} presents the results for each $J$ value in the MOMAB setting. In this case it is clear that the choice of $J$ value has little impact on the algorithms ability to compute the optimal utility. Therefore, given the computational results presented in Figure \ref{fig:bandit_runtime} a lower $J$ value may be preferred.

Although the results presented in Figure \ref{fig:momab_utility} show that the choice of $J$ has little impact on the BTS distributions, it has been shown by Eckles and Kaptein \cite{EcklesKaptein2014} that $J$ values lower than $100$ lead to higher levels of regret in single-objective bandit settings. Therefore, we aim to utilise a $J$ value of around $100$ for DMCTS given the run time for $J$ values of around $100$ are relatively low.  Such values also provide good performance while avoiding the limitations highlighted by Eckles and Kaptein \cite{EcklesKaptein2014}. However, we acknowledge that the $J$ value is problem dependent and certain problems may require a higher $J$.

\subsubsection{Bootstrap Thompson Sampling J Values in MOMDPs}
To evaluate the selection of the J parameter for the BTS distribution has on the performance of DMCTS, we run DMCTS using different J values in a MOMDP. To do so, we have utilised a random MOMDP from the literature \cite{roijers2018multi}. The random MOMDP is configurable based on the requirements of the experiments, where the numbers of states, actions, objectives, timesteps, and possible successor states can be determined a priori. The random MOMDP can then be initialised for each experiment by selecting a consistant seed. We generate a random MOMDP with $20$ states, $2$ actions, and $2$ objectives. The transition function $T(s, a, s')$ is generated using $N = 8$ possible successor states per action, with random probabilities drawn from a uniform distribution \cite{roijers2018multi}. We use the following nonlinear utility function:
\begin{equation}
    u = r_{1}^{2} + r_{2}^{2}.
\end{equation}
\begin{figure}
    \centering
    \includegraphics[width=0.65\textwidth]{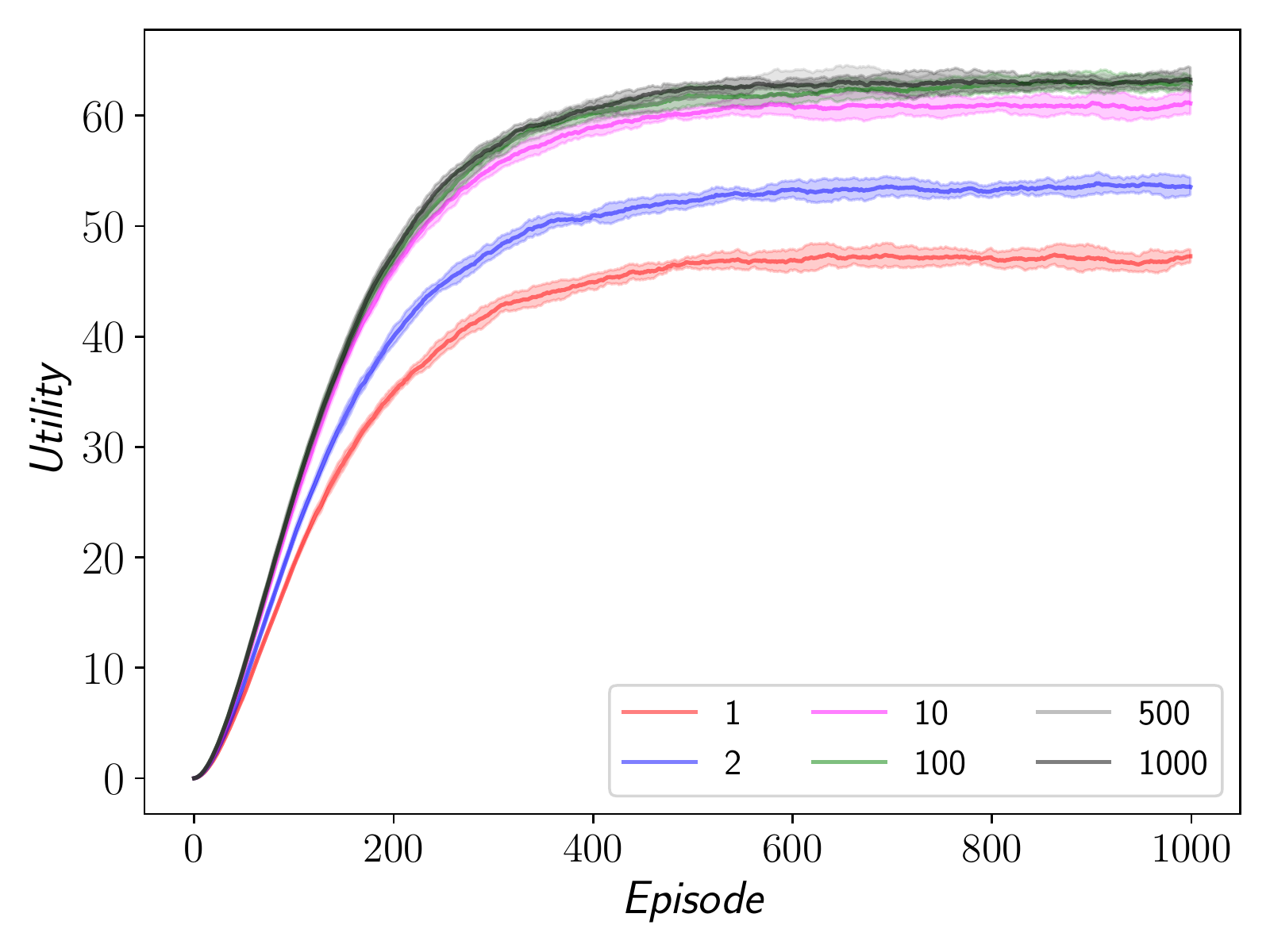}
    \caption{Evaluation of different $J$ values in a random MOMDP with $20$ states, $2$ actions, $2$ objectives and $8$ successor states reachable from each state.}
    \label{fig:j_value_randommomdp}
\end{figure}
We evaluate DMCTS using the following $J$ values of $1, 2, 10, 100, 500$ and $1,000$ for the BTS distributions. Figure \ref{fig:j_value_randommomdp} outlines the results from the random MOMDP. Utilising a $J$ value of $1$ has an impact on performance, given DMCTS with $J$ set to $1$ achieves a lower utility compared to the other parameter settings. As we increase the $J$ value to $2$ we can see that performance begins to improve. However, for a very low $J$ value (J=1 or J=2) DMCTS will select actions greedily and will not explore the environment enough to obtain a good utility. As we increase the $J$ value we can see that the performance increase. Once the $J$ value is set to $10$, DMCTS has a large increase in performance. Similarly, once the $J$ value increases to $100$ we can see even better performance. However, we do not see any more performance increases for DMCTS in the random MOMDP when we set the $J$ value to a higher value. When the $J$ value is set to $500$ or $1,000$ the performance does not increase relative to $J = 100$. However, the computational cost of updates the BTS for higher $J$ values increases. Therefore, it is important to ensure that the $J$ value is set sufficiently high for exploration, while also avoiding $J$ values with a high computational cost. Therefore, it may be important to tune the $J$ value  depending on the evaluation setting.

\subsection{Risk-Aware MDP}
Before testing NLU-MCTS and DMCTS on benchmark problems from the MORL literature, we evaluate both algorithms in a risk-aware problem domain under the ESR criterion. Shen et al.\ \cite{Yun2014} define a Risk-Aware MDP where an agent must decide from a number of stocks in which to invest. The underlying MDP which has $4$ actions (each action is a monetary amount, in Euros, of investment) and $7$ states. At each timestep the agent must select a monetary amount to invest in the stock for a given state. We can invest \texteuro 0, \texteuro 1, \texteuro 2 or \texteuro 3 in a stock at each timestep. Each stock has a probability of making a profit and a probability of making a loss where the agent's return is the action multiplied by the stock price. All remaining implementation details can be found in the work of Shen et al.\ \cite{Yun2014}. In risk-aware decision making a user can be risk seeking, risk averse or risk neutral. A user's preference for risk is described by their individual utility function, which can often be nonlinear. Given risk-based decision making scenarios are ubiquitous in the real world \cite{gerber1998utility} it is important that algorithms can compute policies for risk-aware nonlinear utility functions. Therefore, to highlight the usability of NLU-MCTS and DMCTS in risk-aware decision making scenarios, we evaluate NLU-MCTS and DMCTS using the outlined Risk-Aware MDP. For the Risk-Aware MDP we use a nonlinear risk-seeking and risk-averse utility functions to evaluate the performance of NLU-MCTS and DMCTS.

\subsubsection{Risk-seeking Utility Function}
\label{sec:risk_seeking_experiment}
Firstly, we evaluate DMCTS and NLU-MCTS in the Risk-Aware MDP using the follow risk-seeking utility function:
\begin{equation}
    u(x) = (max(0, x))^2. 
    \label{eqn:risk-seeking}
\end{equation}
In the Risk-Aware MDP, utilising the risk-seeking utility function presented in Section \ref{sec:risk_aware_utility_functions}, would reward the agent with a positive utility when the returns are negative. Therefore, we have used the utility function in Equation \ref{eqn:risk-seeking} to ensure that negative returns are not seen as a positive outcome by the agent. The nonlinear utility function outlined in Equation \ref{eqn:risk-seeking} is risk-seeking given the shape of the utility function is convex.

\begin{figure}
    \centering
    \includegraphics[width=0.65\textwidth]{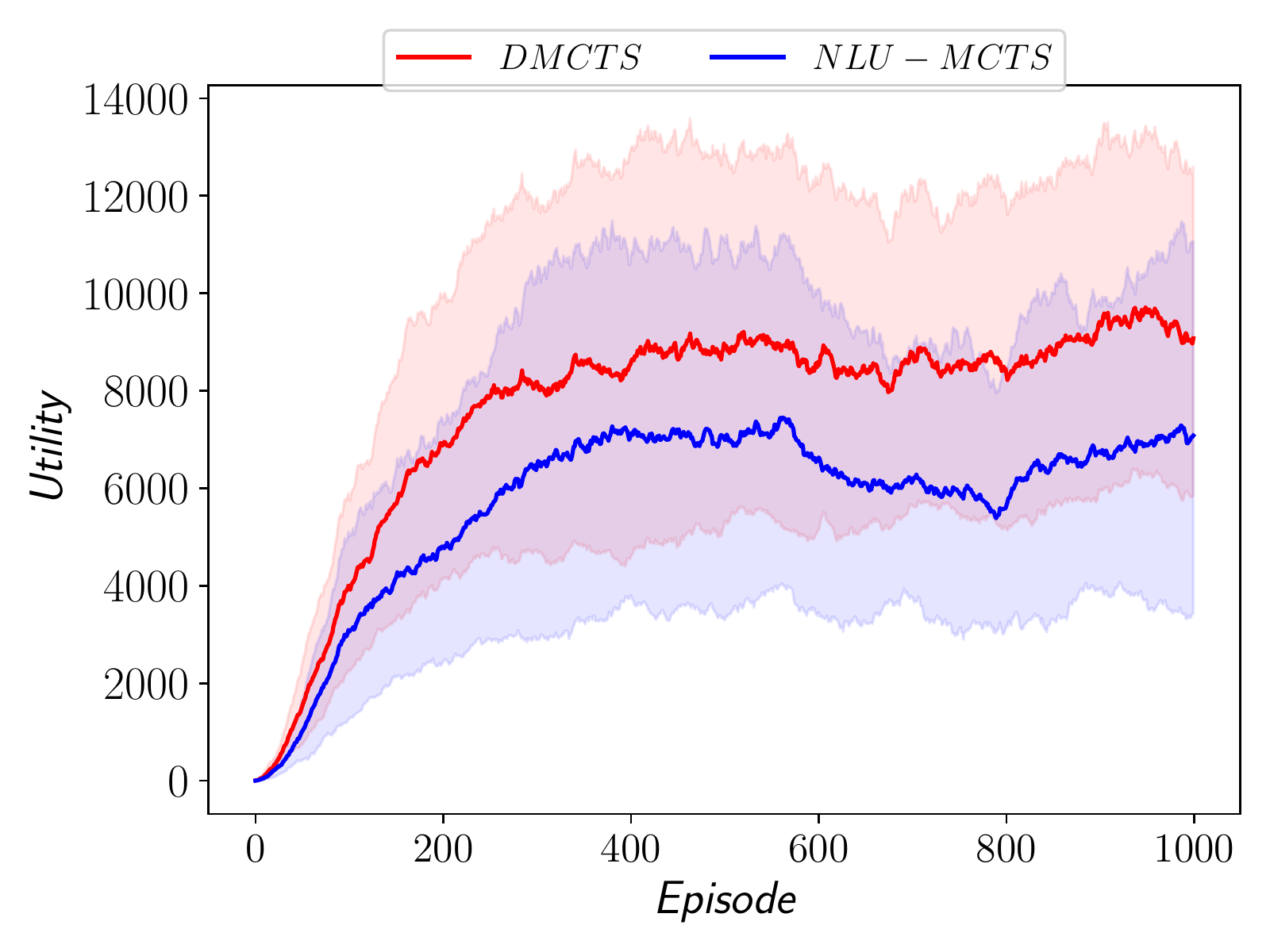}
    \caption{Results from the Risk-Aware MDP environment where DMCTS is evaluated against NLU-MCTS using a risk-seeking utility function. DMCTS achieves a higher utility compared to NLU-MCTS for a risk-seeking utility function.}
    \label{fig:risk_seeking_utility}
\end{figure}

For all experiments in the Risk-Aware MDP with the risk-seeking utility function, the parameter $n_{exec}$ is set to $10$ for each algorithm and each experiment lasts for $1,000$ episodes. For DMCTS we set the number of bootstrap replicates, $J$, for the bootstrap distribution as follows: $J = 500$. For NLU-MCTS we set $C = \sqrt{2}$.

Figure \ref{fig:risk_seeking_utility} describes the experimental results for the risk-seeking utility function in the Risk-Aware MDP. While both algorithms learn good stable policies, DMCTS achieves a higher utility when compared to NLU-MCTS for the risk-seeking utility function.

\subsubsection{Risk-averse Utility Function}
\label{sec:risk_averse_experiment}
Secondly, we evaluate DMCTS and NLU-MCTS using the following risk-averse utility function:
\begin{equation}
    u(x) = x^{\frac{1}{2}}.
    \label{eqn:risk-averse}
\end{equation}
The utility function in Equation \ref{eqn:risk-averse} is risk-averse given the shape of the utility function is concave. 

For all experiments in the Risk-Aware MDP with the risk-averse utility function, the parameter $n_{exec}$ is set to $10$ for each algorithm and each experiment lasts for $1,000$ episodes. For DMCTS we set the number of bootstrap replicates, $J$, for the bootstrap distribution as follows: $J = 500$. For NLU-MCTS we set $C = \sqrt{2}$. In the Risk-Aware MDP the returns can be negative. Therefore, for the risk-averse utility function, we add $150$ the returns because this is the minimum returns that the agent can achieve.

\begin{figure}
    \centering
    \includegraphics[width=0.65\textwidth]{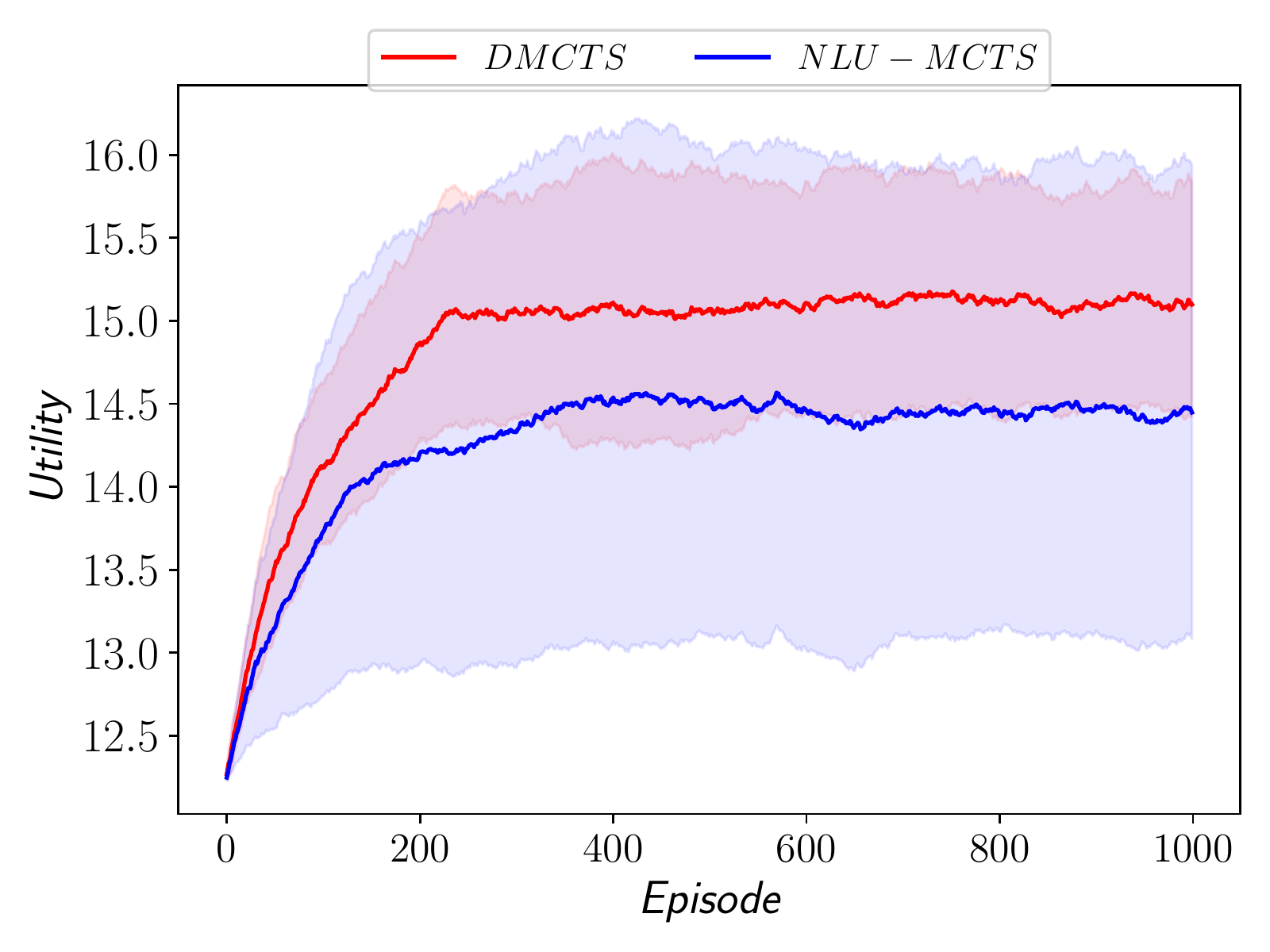}
    \caption{Results from the Risk-Aware MDP environment where DMCTS is evaluated against NLU-MCTS using a risk-averse utility function. DMCTS achieves a higher utility compared to NLU-MCTS for a risk-averse utility function.}
    \label{fig:riskaverse_utility}
\end{figure}

Figure \ref{fig:riskaverse_utility} shows that both NLU-MCTS and DMCTS can compute good policies for the risk-averse utility function. DMCTS achieves a higher utility when compared to NLU-MCTS.

\subsubsection{Discussion of Experimental Results for Risk-Aware MDP}
Both NLU-MCTS and DMCTS learn good policies for the risk-seeking and risk-averse utility functions. However, DMCTS achieves a higher utility in both settings. It is important to note that both the risk-seeking and risk-averse utility function are nonlinear. Therefore for any algorithm to compute good policies the cumulative returns must be taken into consideration. Both NLU-MCTS and DMCTS compute policies based on the expected utility by computing the cumulative returns. If the cumulative returns were not taken into consideration we would expect the utility obtained by both algorithms to be lower, given that nonlinear utility functions do not distribute across the sum of the future and immediate returns. Therefore, for risk-aware settings it is important that the cumulative returns are calculated before the utility function is applied. Taking this approach ensures that an agent can make decisions with knowledge of how future outcomes may affect utility. This is important in risk-aware settings given an agent may only have one opportunity to execute a policy, and having access to the expected utility of a policy ensures sufficient information is available to the agent so utility can effectively be optimised.

While both NLU-MCTS and DMCTS learn good polices for both risk-aware utility functions, DMCTS achieves a higher utility. The key difference between NLU-MCTS and DMCTS is how each algorithm explores during planning. NLU-MCTS utilises UCB while DMCTS uses the Thompson sampling method, Bootstrap Thompson sampling. In the bandit literature Thompson sampling methods have been shown to empirically outperform UCB \cite{chapelle2011thompson}. Thompson sampling selects actions proportional to the probability of an action being optimal \cite{russo2018tutorial}. Therefore, by maintaining an approximate posterior distribution via a bootstrap distribution, and using Thompson sampling to sample from each approximate posterior distribution at each chance node to select actions, DMCTS can exploit the performance gains of Thompson sampling to achieve a higher utility than NLU-MCTS.

\subsection{Multi-Objective MDPs}
\label{sec:exp:mpmdps}
To evaluate NLU-MCTS and DMCTS in multi-objective settings under the ESR criterion, we use a number of problem domains. Firstly, we evaluate NLU-MCTS and DMCTS in the Fishwood problem \cite{roijers2018multi}, given this is one of the very few domains for which ESR results have been published. Secondly, we evaluate NLU-MCTS and DMCTS in the Renewable Energy Dynamic Economic Emissions Dispatch (REDEED) problem domain\footnote{It is important to note, in Section \ref{sec:exp:mpmdps} we evaluate NLU-MCTS and DMCTS (both model-based algorithms) against a number of model-free algorithms. Therefore, to fairly evaluate model-based and model-free approaches, both NLU-MCTS and DMCTS maintain the search tree across episodes. This has the effect of both algorithm need less simulations during experimentation.}.

\subsubsection{Fishwood}
Fishwood is a multi-objective benchmark problem proposed by Roijers et al. \cite{roijers2018multi}. In Fishwood the agent has two states: in the woods or at the river. The goal of the agent is to catch fish and collect wood. The Fishwood environment is parameterised by the probabilities of successfully obtaining fish and wood at these respective states. In this paper we use the following values: at the river the agent has a $0.25$ chance of catching a fish and in the woods the agent has a $0.65$ chance of acquiring wood. For every fish caught, two pieces of wood are required to cook the fish, which results in a utility of $1$. The goal in this setting is to maximise the following nonlinear utility function:
\begin{equation}
    u = \min \left( \texttt{fish}, \left\lfloor \frac{\texttt{wood}}{2} \right\rfloor \right).
\label{eqn:fishwood:utility_function}
\end{equation}

As demonstrated by Roijers et al. \cite{roijers2018multi}, to maximise utility in Fishwood it is essential that both past and future returns are taken into consideration when learning. For example, if there are $5$ timesteps remaining and the agent has received $2$ pieces of wood, the agent should go to the river and try to catch a fish to ensure a utility of $1$ \cite{roijers2018multi}.

We evaluate NLU-MCTS and DMCTS in the Fishwood domain against Expected Utility Policy Gradient (EUPG) \cite{roijers2018multi} and C51 \cite{bellemare2017distributional}. EUPG achieves state-of-the-art results in the Fishwood problem under ESR \cite{roijers2018multi}. C51 \cite{bellemare2017distributional} is a distributional deep reinforcement learning algorithm that achieved state-of-the-art results in the Atari game problem domain.

For C51 the learning parameters were set as follows: $V_{min} = 0$, $V_{max} = 2$, $\epsilon = 0.01$, $\gamma = 1$ and $\alpha = 0.0001$. For DMCTS we set the number of bootstrap replicates, $J$, in the bootstrap distribution as follows: $J = 100$. For NLU-MCTS we set $C = \sqrt{2}$. EUPG is conditioned on the accrued returns and the current timestep, t. We set $n_{exec} = 2$ and run each experiment for $10,000$ episodes where each episode has $13$ timesteps.
\begin{figure}[t]
    \centering
    \includegraphics[width=0.65\textwidth]{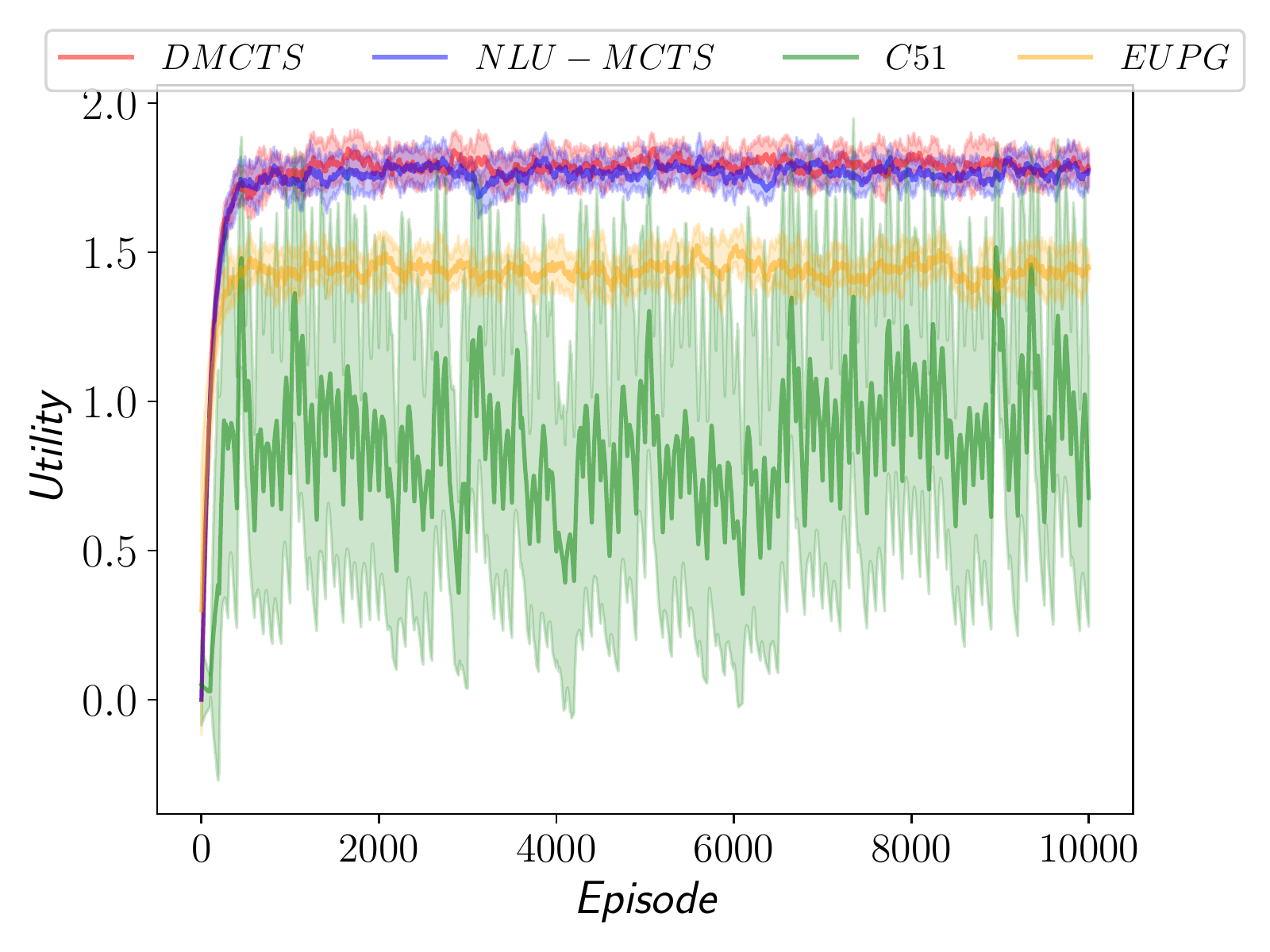}
    \caption{Results from the Fishwood environment where DMCTS achieves state-of-the-art performance in a multi-objective setting over EUPG.}
    \label{fig:fishwood_utility}
\end{figure}

As shown in Figure \ref{fig:fishwood_utility}, the utility for C51 fluctuates throughout experimentation and it fails to learn a consistent policy. Given C51 does not take the accrued returns into consideration during learning the utility function is applied directly to the reward received by the agent. The reward received by an agent in the Fishwood domain can be $[0, 1]$ or $[1, 0]$. Thereby applying the utility function, presented in Equation \ref{eqn:fishwood:utility_function}, to the reward the C51 agent can only receive a utility of $0$. DMCTS, NLU-MCTS and EUPG all take the accrued and future returns into consideration and can learn better policies when compared to C51.

DMCTS, NLU-MCTS and EUPG outperform C51. DMCTS and NLU-MCTS achieve a higher utility when compared to EUPG. All algorithms, except C51, use Monte Carlo simulations of the environment and optimise over the expected utility of the returns of a full episode. Although EUPG uses Monte Carlo simulations of the environment, policy gradient algorithms are sample inefficient. DMCTS and NLU-MCTS are sample efficient given both algorithms utilise the planning phase steps, which has been shown to be sample efficient \cite{Abramson1987,Chang2005}. 

In the Fishwood environment, the agent is not guaranteed to obtain a fish or a piece of wood. For an action in a particular state the agent may need multiple simulations to understand the underlying distribution of the stochastic rewards. Both DMCTS and NLU-MCTS build a search tree, which enables the agent to re-sample the environment at each chance node during learning. However, DMCTS achieves an overall higher utility when compared with NLU-MCTS despite both algorithms utilising repeated sampling at each chance node and Monte Carlo simulations. 

\subsubsection{Renewable Energy Dynamic Economic Emissions Dispatch}
Next, we evaluate NLU-MCTS and DMCTS in a complex problem domain with a large state action space. Renewable Energy DEED (REDEED) is a variation of the traditional DEED problem \cite{BasuDEED}. In REDEED, the power demand for a city must be met over $24$ hours. To supply the city with sufficient power, a number of generators are required. There are $9$ fossil fuel-powered generators, including a slack generator and $1$ generator powered by renewable energy which is generated by a wind turbine. The optimal power output for each generator was derived by Mannion et al.\ \cite{mannion2018reward} and the derived values are used for the both the fossil fuel generators and the renewable energy generator. In this example, Generator 3 is controlled by an agent, Generator 1 is a slack generator and Generator 4 is powered by a wind turbine.

In this setting we imagine a period of $24$ hours and for each hour we receive a weather forecast for a city. For hours $1 - 15$, the weather is predictable and the optimal power values derived by Mannion et al.\ \cite{mannion2018reward} can be used to generate power. From hours $16 - 24$, a storm is forecast for the city. During the storm, both high and low levels of wind are expected and the weather forecast impacts how much power the wind turbine can generate. At each hour during the storm, there is a 0.15 chance the wind turbine will produce $25\%$ less power than optimal, a 0.7 chance the wind turbine will produce optimal power and a 0.15 chance the wind turbine will produce $25\%$ more power than optimal. In the REDEED problem we aim to learn a policy that can ensure the required power is met over the entire day while reducing both the cost and emissions created by all generators.

The goal is to maximise the following nonlinear utility function under the ESR criterion,
\begin{equation}
    \label{eqn:linear_scalarisation}
    R_{+} = \prod_{o=1}^{O} f_{o},
\end{equation}
where $f_{o}$ is the objective function for each objective, $o \in O$ \cite{mannion2018reward,HayesTLO}.

The following equation calculates the local cost for each generator $n$, at each hour $m$:
\begin{equation}
        f_c^L(n,m) = a_n + b_n P_{nm} + c_n (P_{nm})^2 + |d_n sin\{ e_n (P_{n}^{min} - P_{nm}) \}|. 
\end{equation}%
Therefore the global cost for all generators can be defined as:
\begin{equation}
        f_c^G(m) = \sum\limits_{n=1}^N f_c^L(n,m).
\end{equation}
The local emissions for each generator, $n$, at each hour, $m$, is calculated using the following equation :
\begin{equation}
        f_e^L(n,m) = E(a_n + b_n P_{nm} + \gamma_n(P_{nm})^2 + \eta \exp \delta P_{nm}).
\end{equation}%
Therefore the global emissions for all generators can be defined as:
\begin{equation}
        f_e^G(m) = \sum\limits_{n=1}^N f_e^L(n,m).
\end{equation}%
It is important to note the emissions for the generator controlled by the wind turbine are set to $0$.

If the agent exceeds the ramp and power limits a penalty is received. A global penalty function $f_p^G$ is defined to capture the violations of these constraints,
\begin{equation}
    f_{p}^G(m) = \sum\limits_{v=1}^V C(|h_{v} + 1| \delta_v).
    \label{eqn:penaltyHour}
\end{equation}

Along with cost and emissions, the penalty function is an additional objective that will need to be optimised. Some parameters for this problem domain have not been included, all equations and parameters absent from this paper that are required to implement this problem domain can be found in the works of Basu \cite{BasuDEED} and Mannion et al.\ \cite{mannion2018reward}.

To evaluate NLU-MCTS and DMCTS in the REDEED domain, we compare against EUPG and C51. For DMCTS we set the number of bootstrap replicates, $J$, for the bootstrap distribution as follows: $J = 100$. For NLU-MCTS we set $C = \sqrt{2}$. For C51 the learning parameters were set as follows: $V_{min} = -8e^{22}$, $V_{max} = 0$, $\epsilon = 0.01$, $\gamma = 1$ and $\alpha = 0.0001$. For the REDEED problem the agent learns for $10,000$ episodes and $n_{exec} = 2$ for each algorithm.

\begin{figure}
    \centering
    \includegraphics[width=0.65\textwidth]{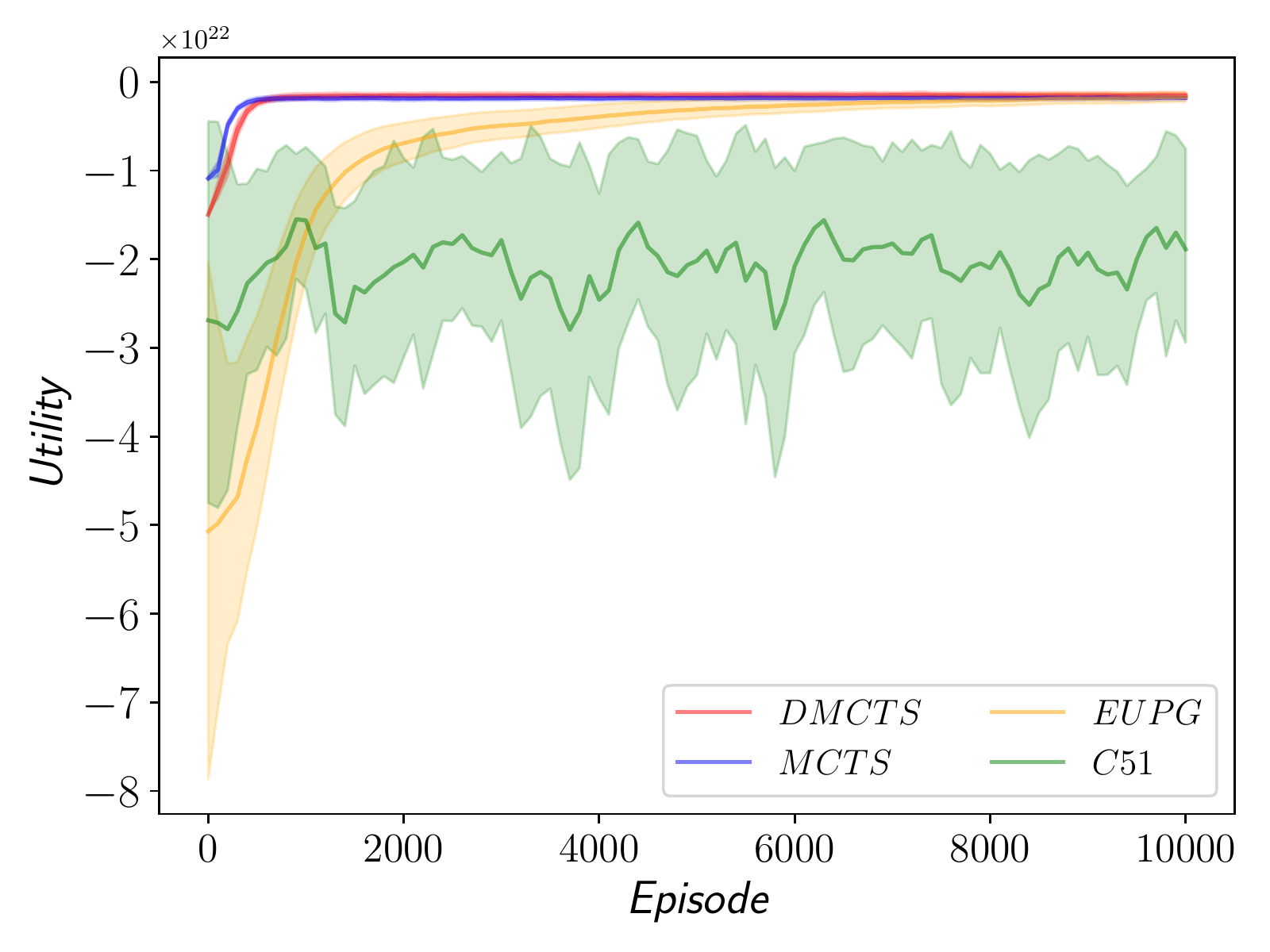}
    \caption{Results from the REDEED environment DMCTS outperforms EUPG, C51 and NLU-MCTS. DMCTS achieves a higher utility compared to other algorithms used throughout experimentation in the REDEED domain under the ESR criterion.}
    \label{fig:deed_utility}
\end{figure}
As seen in Figure \ref{fig:deed_utility}, DMCTS outperforms EUPG, NLU-MCTS and C51 in the REDEED domain. C51 struggles to learn a consistent policy and C51's utility fluctuates throughout experimentation. The hyper-parameters chosen for C51 provide good performance but are difficult to tune. Although the learning speed of EUPG is slow, EUPG achieves a higher utility than C51. 

Both DMCTS and NLU-MCTS learn good policies faster than EUPG. MCTS algorithms are much more sample efficient when compared to policy gradient algorithms like EUPG. Figure \ref{fig:deed_utility} highlights the difference in sample efficiency of DMCTS, NLU-MCTS and EUPG given the differences in the number episodes required for each of the aforementioned algorithms to compute stable policies for the defined nonlinear utility function.

\begin{figure}
    \centering
    \includegraphics[width=0.65\textwidth]{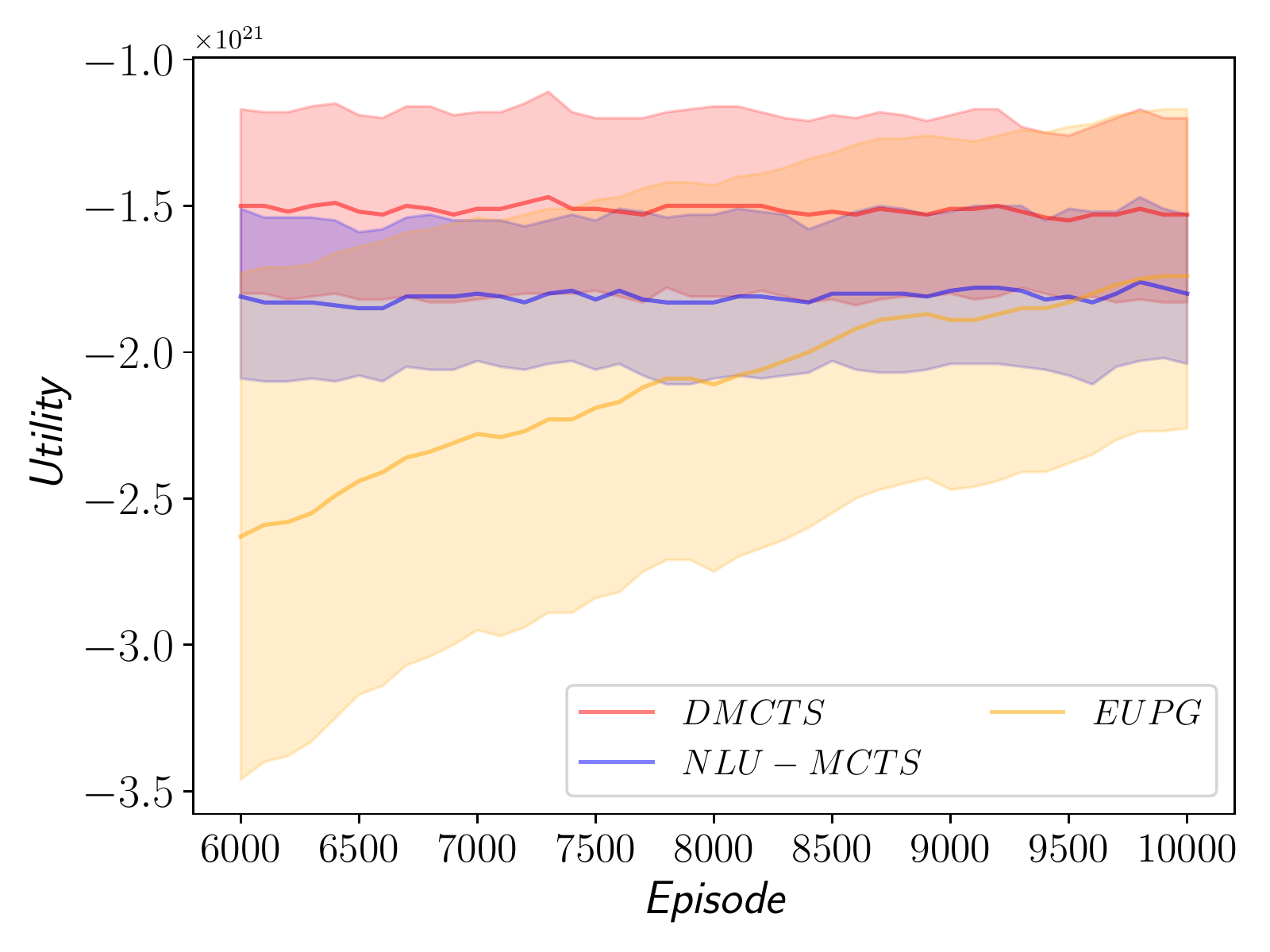}
    \caption{Results from the final 4,000 episodes of the REDEED environment to highlight how DMCTS outperforms NLU-MCTS, EUPG and C51.}
    \label{fig:deed_utility_close}
\end{figure}

DMCTS, NLU-MCTS and EUPG all learn stable policies. However, DMCTS achieves a higher utility when compared to NLU-MCTS and EUPG. DMCTS converges to a policy with an average utility of $-1.54\times 10^{21}$. In comparison, NLU-MCTS converges to a policy with an average utility of $-1.80\times 10^{21}$, while EUPG converges to a stable policy with an average utility of $-1.75\times 10^{21}$. Given the scale of the utility computed throughout REDEED experimentation, it is difficult to see the final difference in utility in Figure \ref{fig:deed_utility}. Therefore, to highlight the difference in utility between DMCTS, NLU-MCTS and EUPG we have plotted the final $4,000$ episodes in Figure \ref{fig:deed_utility_close}. It is important to note, given C51 has performed poorly in the REDEED domain, we have not included C51 in Figure \ref{fig:deed_utility_close}. For the highlighted episodes in Figure \ref{fig:deed_utility_close}, it is clear that DMCTS achieves a higher utility when compared to both NLU-MCTS and EUPG.

The REDEED environment has a large state action space with complex returns. Although C51 has achieved state-of-the-art results in the Atari environment \cite{bellemare2017distributional}, C51 fails to learn any meaningful policy for REDEED. We hypothesise that a reason for poor performance is C51's inability to learn a distribution over the full returns and the level of discretisation of the distribution. The distribution for C51 uses 51 bins to discretise the algorithm's categorical distribution. In the work presented by Bellemare et al.\ \cite{bellemare2017distributional} the number of bins is set to $51$. While this provides good performance, Bellemare et al.\ \cite{bellemare2017distributional} highlight that increasing the number of parameters may lead to increased performance. However, we fix the number of bins to $51$ to remain consistant with the literature, given the potential added performance when increasing the number of bins has not been thoroughly explored. The results presented in this paper for C51 show this parameter setting is sub-optimal in scenarios where the returns are not simple scalars over small ranges. 
The results present in Figure \ref{fig:deed_utility} show that C51 struggles to scale to large problem domains with complex returns over large ranges. 

\subsubsection{Discussion of Experimental Results for Multi-Objective MDPs}
In multi-objective settings both NLU-MCTS and DMCTS learn good policies for the specified nonlinear utility functions. Similarly to the Risk-Aware MDP, applying the utility function to the cumulative returns (rather than just the expected future return) ensures that both NLU-MCTS and DMCTS can learn good policies. It is clear from the performance of C51 that applying the utility function to the cumulative returns is crucial for good performance in multi-objective settings when the utility function is nonlinear.

As previously highlighted, the difference between NLU-MCTS and DMCTS is the method used to explore during planning. NLU-MCTS uses UCB to determine which action to take during planning. UCB selects actions deterministically based on the expected utility and an exploration bonus \cite{russo2014learning,auer2002using}. In contrast DMCTS selects actions using Thompson sampling, which stochastically samples from the underlying approximate posterior distribution (BTS distribution) and selects the action proportional to the probability of the action being optimal \cite{chapelle2011thompson,EcklesKaptein2014}. In the bandit literature Thompson sampling has been shown to empirically outperform UCB \cite{chapelle2011thompson,russo2018tutorial,russo2014learning}. Monte Carlo tree search methods utilise independent nodes, therefore we can consider each node itself to be a bandit. In this case, we expect Thompson sampling methods to also outperform UCB in sequential settings. Our findings for sequential settings in risk-aware and multi-objective settings are consistent with prior bandit literature that suggests that Thompson sampling can outperform UCB \cite{chapelle2011thompson}.

Additionally, UCB makes highly pessimistic assumptions regarding the underlying reward/return distributions, in order to guarantee a bound on the regret of its action selection procedure \cite{russo2014learning}. For known parametric distributions, tighter bounds have been proven using tighter upper confidence bounds (e.g. for Gaussian reward distributions \cite{russo2014learning}). However, doing something similar in our setting isn't opportune, because even if the return distributions are nicely parametric, the nonlinear transformation resulting from the application of the utility function would no longer allow for a closed-form distributions \cite{russo2014learning}. As such, we are either stuck with highly pessimistic assumptions (and therefore suboptimal performance) or we need to have a different method. Bootstrap distributions and the resulting Bootstrap Thompson sampling algorithm for action selection is able to approximate, and effectively exploit knowledge about the utility distributions, regardless of the shape of this underlying distribution.

\subsection{Nonlinear Utility Functions}
During experimentation DMCTS has been evaluated using previously defined utility functions for each experimental benchmark. To show that DMCTS can learn a good policy for any nonlinear utility function we have evaluated DMCTS in the Fishwood problem domain using four nonlinear utility functions under the ESR criterion. The following nonlinear utility functions are used to evaluate DMCTS in the Fishwood domain:
\begin{equation*}
    u_{1} = max(\frac{r_{1}}{2}, \frac{r_{2}}{2} ) ,
\end{equation*}
\begin{equation*}
    u_{2} = \frac{r_{1}}{2} + r_{2}^{4} ,
\end{equation*}
\begin{equation*}
    u_{3} = min(\frac{r_{1}}{2}, \frac{r_{2}}{4} ) ,
\end{equation*}
\begin{equation*}
    u_{4} = r_{1}^{2} + r_{2}^{2},
\end{equation*}
where $r_{1}$ is the returns received for the fish objective and $r_{2}$ is the returns received for the wood objective.

For this demonstration, we set $n_{exec} = 2$ and each experiment lasts $10,000$ episodes. For DMCTS we set the number of bootstrap replicates, $J$, for the bootstrap distribution as follows: $J = 100$.

\begin{figure}
    \centering
    \includegraphics[width=0.65\textwidth]{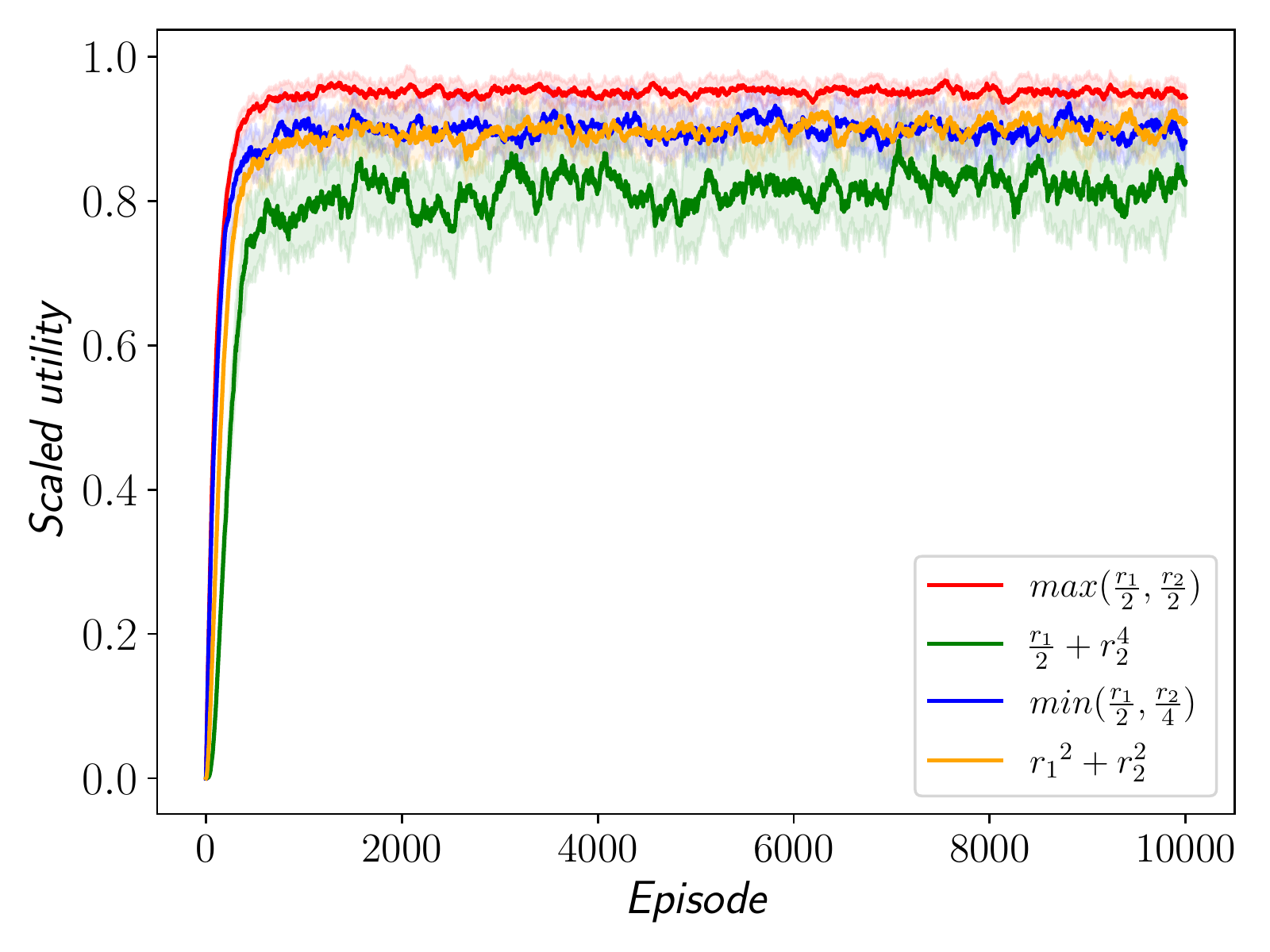}
    \caption{Results from the Fishwood environment where DMCTS is evaluated against multiple nonlinear utility functions.}
    \label{fig:fishwood_nonlinear_utility}
\end{figure}

In Figure \ref{fig:fishwood_nonlinear_utility}, for each utility function we have scaled the utility between $0$ and $1$. For the scaled utility, $1$ represents the maximum utility and $0$ represents the minimum utility obtained by DMCTS. We have scaled the utility to show the performance of DMCTS for each utility function on a single plot.

Figure \ref{fig:fishwood_nonlinear_utility} outlines the performance of DMCTS when optimising for each nonlinear utility function. It is clear from Figure \ref{fig:fishwood_nonlinear_utility} that DMCTS converges to a good policy for each utility function. Therefore, DMCTS can learn a good policy for each of the outlined nonlinear utility functions and is not limited to the utility functions associated with predefined benchmark problems. The ability of DMCTS to learn a good policy for a range of nonlinear utility showcases how DMCTS could potentially be used in real-world scenarios, where different decision makers may have very different nonlinear utility functions for the same problem.

\section{Related Work}
Many risk-aware RL approaches seek to learn policies to maximise the expected return. Some research in this area focuses on learning policies which maximise the expected exponential utility \cite{Moldovan2012}. Other approaches take the weighted sum of the return and risk into consideration when learning policies \cite{Gosavi2009,Geibel2005}. Although most risk-aware RL approaches aim to maximise the expected utility, they often do not take into consideration the utility of the return of a full episode. It is also important to note that little research exists where decisions are made based on a learned distribution over the expected returns \cite{Morimura2010a,Morimura2010b} for risk-aware RL. 

Many MCTS methods have developed for situations involving reward uncertainty. For example, Tesauro et al. \cite{tesauro2012bayesian} take a Bayesian approach to UCT with Gaussian approximation. Their method backpropagates probability distributions over rewards. To select actions Tesauro et al. \cite{tesauro2012bayesian} use UCB1 with taking the distributions into consideration. Cazenave and Saffidine \cite{cazenave2010score} define a MCTS algorithm that takes into account the bounds on the possible values of a node to select nodes for exploration. They apply their algorithm to problems that have more than two outcomes and show that taking the bounds into consideration can increase performance. Kaufmann and Koolen \cite{kaufmann2017monte}, and Huang et al. \cite{huang2017mcts} also have developed MCTS algorithms which can compute policies for settings with reward uncertainty.

As previously highlighted, the majority of RL research focuses on the SER criterion. Multi-objective MCTS (MOMCTS) \cite{Sebag2012} was shown to be able to learn a coverage set under SER. However, MOMCTS can only learn a coverage set in deterministic environments. Convex Hull MCTS \cite{Painter2020} is able to learn the convex hull of the Pareto front but focuses solely on linear utility functions. A number of other multi-objective MCTS methods exist \cite{Perez2013,Jongmin2018,Perez2015}, but no method has previously been shown to learn the Pareto front for both deterministic and stochastic environments for any unknown utility function. In contrast to the SER criterion, no method exists that can learn a set of optimal policies under the ESR criterion in sequential settings. Hayes et al. \cite{hayes2021expected} compute a set of ESR non-dominated return distributions, known as the ESR set, in a multi-objective multi-armed bandit setting. However, the method proposed by Hayes et al. \cite{hayes2021expected} cannot be applied to sequential decision making problems.
An interesting opportunity for future work is the possibility of building on the methods of Wang and Sebag \cite{Sebag2012} and Painter et al.\ \cite{Painter2020} to extend DMCTS to learn the optimal coverage set under both SER and ESR for any unknown utility function.

Zhang et al. \cite{zhang2021distributional} compute a multi-variate distribution over the returns for RL settings. While this work considers reward vectors, we believe this algorithm will suffer from similar limitations to traditional RL algorithms when applied to nonlinear utility functions. The method proposed by Zhang et al. \cite{zhang2021distributional} does not take the accrued returns into consideration. Therefore we believe that such an approach would fail to achieve a high utility \cite{roijers2018multi}. However, this method could be an interesting starting point for developing model-free multi-objective distributional RL algorithms.

A key argument in this paper is that the expected utility of the future returns under ESR must be replaced with a posterior distribution over the expected utility of the returns. Bai et al.\ \cite{Bai2014} extend MCTS to maintain a distribution at each node using Thompson Sampling as an exploration strategy. However, the work presented in this paper is significantly different. In their work, Bai et al.\ \cite{Bai2014} do not learn a posterior distribution over the expected utility of the return, apply their work to multi-objective settings, or incorporate the accrued returns as part of their algorithm. It is also important to note the C51 algorithm proposed by Bellemare et al.\ \cite{bellemare2017distributional} achieves state-of-the-art performance in single-objective settings and learns a distribution over the future returns. Abdolmaleki et al.\ \cite{Abdolmaleki2020} learn a distribution over actions based on constraints set per objective. This approach ignores the utility-based approach \cite{roijers2013survey} and uses constraints set by the user to learn a coverage set of policies where the value of constraints is dependent on the scale of the objectives. Abdolmaleki et al. \cite{Abdolmaleki2020} claim setting the constraints for this algorithm is a more intuitive approach when compared to setting weights for a linear utility function. We theorise that if the user's utility function is nonlinear, this approach would fail to learn a coverage set.

\section{Conclusion \& Future Work}
In this paper we propose a novel Monte Carlo tree search algorithm that can compute good policies for nonlinear utility functions (NLU-MCTS). We then extend NLU-MCTS, to define a new distributional Monte Carlo tree search (DMCTS) algorithm. Both NLU-MCTS and DMCTS are able to learn good policies in MORL settings, under the ESR criterion for nonlinear utility functions in problem domains with stochastic rewards. DMCTS replaces the expected utility of the future returns with a bootstrap distribution over the utility of the returns, and achieves state-of-the-art performance in MORL domains under the ESR criterion. We achieve this by using a bootstrap distribution as an approximate posterior over the expected utility of the returns of the episode. It is our hope that this paper will inspire further work on algorithms that replace the expected returns with a distribution over the expected utility of the returns for risk-aware and ESR settings.

Although DMCTS achieves state-of-the-art performance under the ESR criterion, as the size of the problem domain increases we expect the bootstrap distribution may encounter limitations. In order to apply DMCTS to real-world problem domains like, \cite{ABRAMS2021covid}, distributions like Dirichlet \cite{olkin1964multivariate} may be better suited to high dimensional state spaces especially when dealing with multi-variate scenarios when learning policies for the SER criterion. 

We also aim to extended both NLU-MCTS and DMCTS to learn policies in multi-objective environments with continuous state spaces. For example, Abels et al. \cite{abels2019dynamic} define a multi-objective benchmark problem known as Minecart that has a continuous state space. In the future we plan to extend DMCTS to learn policies in problem domains with continuous action spaces \cite{xu2020prediction}. 

DMCTS initialises the $\alpha$ and $\beta$ values for each bootstrap replicate to $1$. However, not much is known about the impact of such an initialisation. It is possible that varying this value could increase performance. Therefore, we plan to investigate the potential performance gains possible from altering this parameter on initialisation.


In the work of Martin et al. \cite{martin2020stochastically} stochastic dominance \cite{wolfstetter_1999,levy1992stochastic,hanoch1969} is used to determine optimal actions by comparing learned categorical distributions over the returns. Martin et al. \cite{martin2020stochastically} demonstrate good performance in risk-based scenarios. Hayes et al. \cite{hayes2021esr_set,hayes2021expected} present ESR dominance as a dominance criteria for distributional multi-objective decision making under the ESR criterion. In future work we aim to utilise stochastic dominance with DMCTS for single objective risk-based problems. We also plan to extend DMCTS to utilise ESR dominance as a dominance criterion for multi-objective decision making problems under the ESR criterion.

In this paper, the utility function is known a priori. In different MORL scenarios, the utility function can be unknown at the time of learning or planning \cite{roijers2013survey,radulescu2020survey,hayes2021practical}. In these scenarios, an algorithm must recover a coverage set of optimal policies. Under the SER criterion many methods have been developed that compute sets of optimal policies \cite{white1982multi,wiering2014model,parisi2016multi,bryce2007probabilistic,yang2019generalized}. For example, multi-objective MCTS \cite{Sebag2012} can learn a coverage set for deterministic environments under SER. Currently, no method exists that can compute a set of optimal policies for the ESR criterion by interacting with the environment in a sequential setting. In future work, we aim to extend our DMCTS algorithm to be able to learn coverage sets for unknown utility functions under the ESR criterion and the SER criterion for stochastic environments.

\section*{Acknowledgements}
Conor F.\ Hayes is funded by the University of Galway Hardiman Scholarship. This research was supported by funding from the Flemish Government under the ``Onderzoeksprogramma Artificiële Intelligentie (AI) Vlaanderen'' program. 

%
\section*{Conflict of interest}
The authors declare that they have no conflict of interest.

\bibliography{references.bib}
\bibliographystyle{spmpsci}

\end{document}